\documentclass{article}

\usepackage{enumitem}
\usepackage[preprint]{neurips_2026}

\usepackage[utf8]{inputenc} 
\usepackage[T1]{fontenc}    
\usepackage{url}            
\usepackage{booktabs}       
\usepackage{amsfonts}       
\usepackage{nicefrac}       
\usepackage{xcolor}         
\usepackage{graphicx}
\usepackage{booktabs}   
\usepackage{multirow}   
\usepackage{amsmath}
\usepackage{wrapfig}
\usepackage{float}  
\usepackage{algorithm}
\usepackage{algorithmic}
\usepackage{makecell}
\usepackage{marvosym}
\makeatletter
\@ifpackageloaded{hyperref}{
  \protected\def\showhyphens#1{\HyOrg@showhyphens{#1}}
}{}
\makeatother

\usepackage[dvipsnames]{xcolor}
\usepackage{graphicx}
\usepackage{subcaption}
\usepackage{booktabs}
\usepackage{amssymb}
\usepackage{colortbl}
\usepackage[
  pagebackref=true,
  breaklinks=true,
  bookmarks=true,
  colorlinks,
]{hyperref}
\usepackage{pgfplots}
\pgfplotsset{compat=1.18}

\usepackage{tabularx}
\usepackage{booktabs}
\usepackage{ragged2e}
\newcolumntype{L}{>{\RaggedRight\arraybackslash}X}
\patchcmd{\IEEEbiography}{8pt}{0pt}{}{}

\title{MV-WAM: Manifold-Aware World Action Model with Value Augmentation}

\usepackage{xcolor} 

\author{
    \textbf{Jintao Chen$^{1,2}$}
    \thanks{Equal Contribution.
    $^\dagger$ Project Leader.
    $^{1}$State Key Laboratory of Multimedia Information Processing, School of Computer Science, Peking University.
    $^{2}$Beijing Innovation Center of Humanoid Robotics.
    \textsuperscript{\Letter} Corresponding author.}~~, 
    \textbf{Peidong Jia$^{1,2,*,\dagger}$}, 
    \textbf{Qingpo Wuwu$^{1,2,*}$}, 
    \textbf{Jiaming Liu$^{2}$},
    \textbf{Mengfei Du$^{2}$},
    \vspace{0.099em} \\
    \textbf{Chun-Kai Fan$^{1,2}$},
    \textbf{Xiaowei Chi$^{1,2}$},
    \textbf{Hao Chen$^{2}$},
    \textbf{Chengyu Bai$^{1,2}$},
    \textbf{Zezhong Qian$^{1,2}$},
    \vspace{0.099em} \\
    \textbf{Hao Wang$^{1,2}$},
    \textbf{Jiajun Cao$^{1,2}$},
    \vspace{0.099em} 
    \textbf{Weishi Mi$^{2}$},
    \textbf{Xiaozhu Ju$^{2}$,} 
    \textbf{Jian Tang$^{2}$,} 
    \textbf{Shanghang Zhang$^{1},$\textsuperscript{\Letter}}
}

\begin{document}

\maketitle

\begin{figure}[H]
    \centering
    \vspace{-0.7cm}
    \includegraphics[width=\textwidth]{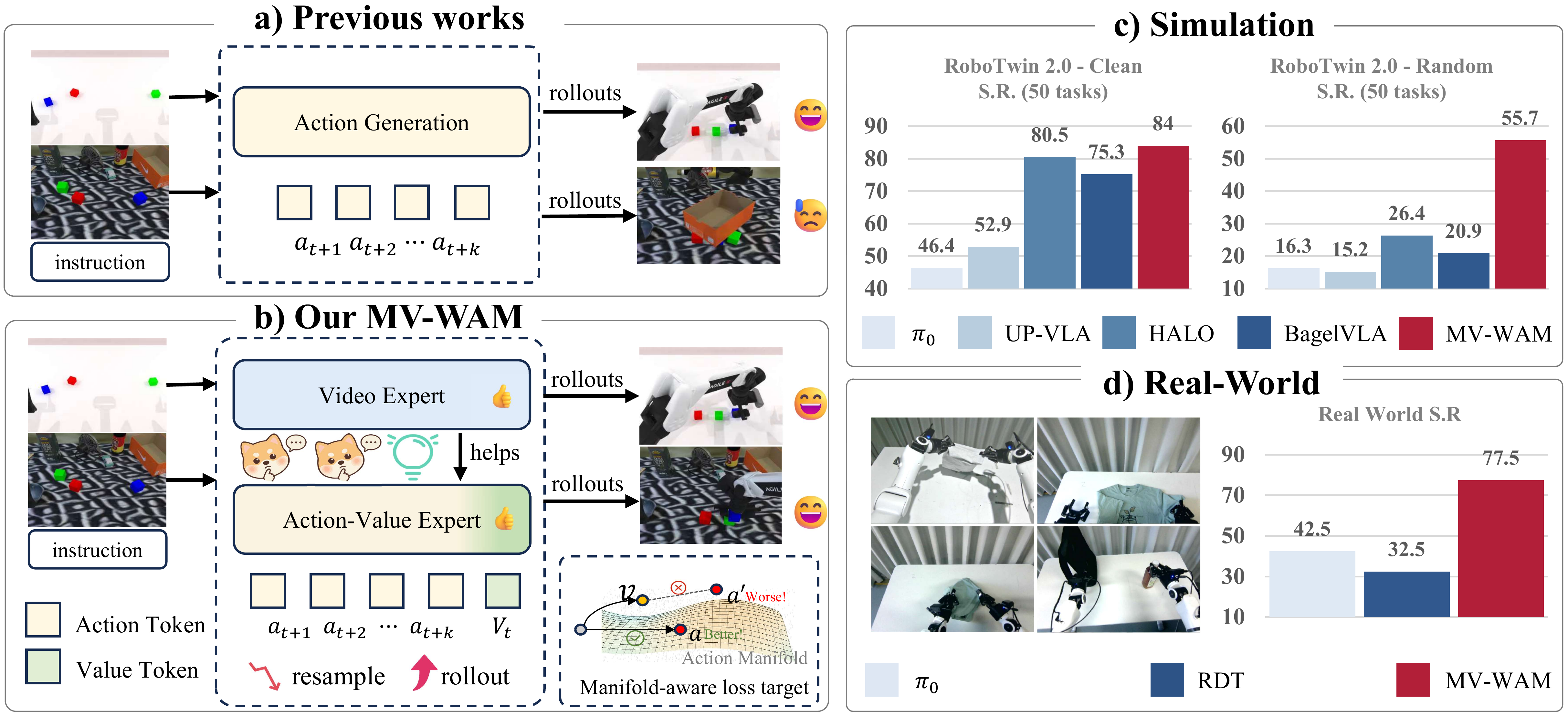}
    \caption{\textbf{Overview of MV-WAM.} We introduce asymmetric video-action experts with a causal mask to condition actions on visual dynamics. By coupling world modeling, action prediction, and progress-value estimation, it achieves strong performance in (c) simulation and (d) real-world tasks.}
    \label{fig:teaser}
    \vspace{-0.35cm}
\end{figure}
\begin{abstract}

Achieving robust and generalizable manipulation across diverse environments remains a fundamental challenge in embodied robotics. Recent world action models achieve strong in-domain performance, yet their 
gains do not extend proportionally to out-of-distribution scenarios. We attribute this to a structural mismatch between visual and action modalities, whose intrinsically heterogeneous manifolds cause joint optimization to disproportionately degrade action robustness under distribution shift. To address this, we propose \textbf{MV-WAM}, a novel end-to-end framework that jointly models visual prediction, action generation, and value estimation designed to effectively leverage video priors during both training and inference for enhanced action generalization. Key to this unification is a cross-modality causal mask that hierarchically grounds actions in predicted video frames and value function tokens in both modalities. To further narrow the generalization gap, MV-WAM adopts a manifold-aware optimization scheme that explicitly accounts for the structural heterogeneity across modalities. Finally, MV-WAM introduces a progress-value regulation mechanism that estimates task completion and detects misalignment between predicted frames and generated actions, enabling the policy to autonomously identify execution deviations and recover through value-guided rollback. On the RoboTwin simulation, MV-WAM 
achieves a \textbf{55.7\%} mean success rate on random scenarios without any randomized action supervision, outperforming the strongest baseline by \textbf{29.3\%}. MV-WAM achieves a \textbf{77.5\%} mean success rate across four real-world tasks of varying difficulty on a dual-arm robot. Our results demonstrate that manifold-aware cross-modal alignment is essential for robust policy generalization, offering a path toward deployable robotic manipulation.

\end{abstract}



\section{Introduction}
Generalization is fundamental to deploying robotic manipulation policies in the real world~\cite{gao2025taxonomy}. Beyond the training distribution, policies must achieve visual generalization to unseen textures, lighting, backgrounds, and object configurations. Recent vision-language-action (VLA) models~\cite{brohan2022rt, driess2023palm, 2024_9_05_OpenVLA}, built upon large-scale vision-language pretraining~\cite{radford2021learning, alayrac2022flamingo, achiam2023gpt, radford2021clip}, have substantially advanced manipulation policies in semantic and task-level generalization. However, this progress has not extended to visual and environmental generalization~\cite{gao2025taxonomy, shi2023groot},
as policies trained on limited demonstrations cannot capture the visual diversity of real-world deployment. World Action Models (WAMs) extend the VLA paradigm by further leveraging large-scale action-free video data which captures far richer visual diversity than demonstration data alone~\cite{hafner2019dream, ha2018world, ye2026world_wans}, broadening visual generalization to regimes that curated demonstrations cannot reach. Yet realizing this promise has proven non-trivial.

Existing WAMs differ chiefly in how they integrate video prior into policies~\cite{bi2025motus,li2026causal,ye2026world_wans,hu2025astranav,yuan2026fastwam}. 
The most loose coupling is achieved by \textit{decoupled} approaches, where a separate video model and policy interact only through generated visual predictions or latent representations~\cite{du2023unipi, yang2024unisim, hu2024vpp, tian2024seer}. \textit{Fully unified} approaches eliminate this distinction altogether, tokenizing visual and action signals into a single stream processed by shared transformer parameters~\cite{cheang2024gr2, zhu2024irasim}. Approaches based on \textit{Mixture-of-Transformers (MoTs)}~\cite{liang2024mixture_of_transformers} go further still, introducing modality-specific experts while sharing global attention, and currently achieve the strongest in-domain performance among WAM designs.
Across these designs, all report consistent improvements over non-WAM baselines on in-domain and out-of-distribution (OOD) scenes alike. Despite these advances, as shown in Appendix~\ref{app:manifold}, in-domain performance steadily climbs while OOD gains lag behind, leaving a wide and in some cases growing generalization gap. Regardless of the architecture, the gains from video prior seem to flow disproportionately into visual prediction rather than action robustness, which is precisely what WAMs were designed to improve. Motivated by these observations, we aim to explore the following question: \textbf{\textit{Why does deeper video-action coupling not translate to greater action generalization?}}

We trace this gap to a manifold mismatch deeply embedded in current WAM 
architectures. Visual observations are high-dimensional and densely structured, 
governed by spatial and photometric regularities, whereas robot actions are 
low-dimensional, temporally structured, and precision-critical over control 
dimensions~\cite{brohan2022rt, cheang2024gr2}. These two modalities thus inhabit 
intrinsically heterogeneous representational manifolds~\cite{yang2026abot}. Joint optimization over 
incompatible manifolds inevitably compromises action learning~\cite{yu2020gradient}, 
leaving action representations under-trained and brittle. This is further aggravated 
by the sheer volume of visual tokens, which naturally dominates the shared objective. 
This misalignment is masked in-domain by strong cross-modal correlations, but under 
distribution shift, visual perturbations propagate through shared parameters and 
degrade action prediction~\cite{gao2025taxonomy}. This structural mismatch is 
architecture-agnostic, explaining why neither decoupled, fully unified, nor MoT-based 
designs have succeeded in closing the generalization gap.

As illustrated in Fig.~\ref{fig:teaser}, we propose \textbf{MV-WAM} (Manifold-Aware World Action Model with Value Augmentation), a novel framework 
that jointly generates future visual imaginations, robot actions, and value estimation within a unified architecture. At its core, MV-WAM adopts a 
MoTs backbone~\cite{liang2024mixture_of_transformers} 
in which each transformer layer hosts two modality-specific experts, one 
processing visual tokens and the other processing action-value tokens. 
Modality-specific experts share a global attention space while maintaining separate parameters. 
The visual expert is initialized from a large-scale pretrained video 
generative model~\cite{chi2025wow}, allowing MV-WAM to inherit a rich visual 
prior over physical dynamics, object interactions, and scene geometry that 
demonstration data alone cannot provide. Built on this framework, three principled components collectively enable action-free video to serve as a direct source of policy generalization. First, a \textit{modality-adaptive optimization} scheme extends the plain MoTs  into a manifold-aware multi-objective MoTs formulation, with each modality supervised according to the intrinsic structure of its target manifold. This eliminates cross-modal optimization interference and alleviates gradient imbalance, preventing each modality's learning signal from corrupting the other.
Second, a \textit{cross-modality causal mask} hierarchically routes information across modalities, grounding each action token in its corresponding predicted visual frame and each value token in both visual and action contexts, ensuring that action generation is always conditioned on the predicted visual future while preserving modality-specific structure. Third, a \textit{progress-valued regulation} mechanism uses value 
tokens, trained with Monte Carlo task returns, to jointly estimate task 
progress and detect visual-action misalignment during execution. During online execution, when 
predicted values drop below a learned threshold, MV-WAM triggers a 
value-guided rollback, enabling the policy to autonomously identify and 
recover from execution deviations.

We evaluate MV-WAM extensively across simulation and real-world settings. On the RoboTwin 2.0~\cite{chen2025robotwin} benchmark, 
covering 50 manipulation tasks across both clean scenes 
and randomized scenes with perturbed lighting, backgrounds, 
and object configurations. MV-WAM achieves state-of-the-art performance under 
both settings, reaching a 55.7\% success rate on unseen scenarios 
and outperforming the strongest baseline by 29.3\%
while matching its in-domain performance. On real-world dual-arm manipulation, MV-WAM further achieves 77.5\% mean success rate across four tasks of varying difficulty, where baseline methods fail to generalize to physical deployment. Notably, MV-WAM achieves these gains with significantly fewer parameters than competing WAMs, suggesting that manifold-aware design rather than model scale is the key driver of generalization. Ablation studies further confirm 
that each architectural component contributes substantively to the OOD 
performance.

Our contributions are summarized as follows.
\textbf{(1) MV-WAM framework.} We propose MV-WAM, a unified world action 
model built on MoTs that jointly generates visual predictions, robot actions, 
and value estimates, with a manifold-aware multi-objective formulation and 
a cross-modality causal mask that ground action generation in predicted 
visual futures while respecting the intrinsic geometry of each modality.
\textbf{(2) Progress-valued regulation.} We introduce a progress-valued 
regulation mechanism that estimates task completion and detects visual-action 
misalignment, enabling autonomous deviation 
detection and value-guided recovery without human intervention.
\textbf{(3) State-of-the-art generalization.} Extensive experiments on 
simulated and real-world manipulation tasks demonstrate state-of-the-art 
performance across in-domain, zero-shot, and few-shot settings. 
\section{Problem Definition and Analysis}

\subsection{Problem Formulation}

We model robotic manipulation as a Partially Observable Markov Decision Process (POMDP)~\cite{kaelbling1998planning_pomdp} $\langle \mathcal{S}, \mathcal{O}, \mathcal{A}, \mathcal{T}, \mathcal{L} \rangle$, where $\mathcal{S}$ denotes the set of environment states, $\mathcal{O}$ the set of visual observations, $\mathcal{A}$ the action space, $\mathcal{T}: \mathcal{S} \times \mathcal{A} \rightarrow \Delta(\mathcal{S})$ the stochastic transition kernel, and $\mathcal{L}$ the set of natural language task instructions. At each timestep $t$, the robot receives a task instruction $l \in \mathcal{L}$, perceives a history of visual observations $o_{t-h_O:t} \in \mathcal{O}^{h_{O+1}}$ as a partial observation of the underlying state $s_t \in \mathcal{S}$, and executes an action chunk $\mathbf{a}_{t:t+h_A} \in \mathcal{A}^{h_{A+1}} \subset \mathbb{R}^{h_{A+1} \times d_a}$, where $h_{A+1}$ is the action horizon and $d_a$ is the action dimensionality. Upon completion of each action chunk, the robot observes the resulting environmental state and decides to either  generate the next step or re-plan.

To solve the above POMDP problem, existing manipulation policies learn a parameterized mapping $\pi_\theta: \mathcal{O}^{h_{O+1}} \times \mathcal{L} \rightarrow \mathcal{A}$. As a common foundational scheme, Standard VLA models learn $\pi_\theta$ that maps  historical observations and language instructions to action chunks, trained exclusively on action-paired demonstration data. WAMs extend this by incorporating abundant action-free video data as a visual prior, enabling $\pi_\theta$ to learn visual dynamics beyond paired demonstrations alone.


\subsection{Preliminaries}

We instantiate the action space $\mathcal{A}$ following the unified action design of RDT-1B~\cite{liu2024rdt}. Each action $\boldsymbol{a}_t \in \mathbb{R}^{128}$ encodes both joint-level and end-effector control. Taking dual-arm robot as an example, this comprises 14-DoF relative joint offsets across the two arms and 2-DoF independent gripper states, standardized into the 128-dimensional representation for flexible and compatible manipulation.

\paragraph{Flow Matching}

We adopt conditional flow matching~\cite{lipman2022flow} as the generative backbone for both visual and action generation. Taking the action generation branch as an example, given a clean target $\boldsymbol{a}_0 \in \mathcal{A}$, Gaussian noise $\boldsymbol{a}_1 \sim \mathcal{N}(\boldsymbol{0},\boldsymbol{I})$ and diffusion timestep $\tau \in [0,1]$, flow matching defines a linear interpolation $\boldsymbol{a}_\tau = (1-\tau)\boldsymbol{a}_0 + \tau\boldsymbol{a}_1$ and trains the policy network $\pi_\theta$ to predict the vector field $\boldsymbol{v}_\theta(\boldsymbol{a}_\tau, \tau, \mathcal{C})$ conditioned on multi-modal context $\mathcal{C} = (\mathcal{O}^{h_O}, \mathcal{L}, \boldsymbol{s}_t)$ via:
\begin{equation}
    \mathcal{L}_{\text{FM}} = \mathbb{E}_{\tau,\boldsymbol{a}_0,\boldsymbol{a}_1}\Big\|\boldsymbol{v}_\theta(\boldsymbol{a}_\tau,\tau,\mathcal{C}) - (\boldsymbol{a}_1 - \boldsymbol{a}_0)\Big\|_2^2.
\end{equation}
At inference, the clean action chunk $\boldsymbol{a}_0$ is recovered by integrating $\boldsymbol{v}_\theta$ from $\boldsymbol{a}_1$ back to $\tau=0$. Both the action branch and the video branch share the identical flow matching pipeline.

\subsection{Manifold Analysis}

We empirically verify that the generalization gap stems from an intrinsic manifold asymmetry between modalities. Existing WAMs exhibit a larger absolute clean-to-random gap than non-unified baselines (54.0--54.8 vs. 20.8--30.1 percentage drop), and t-SNE analysis confirms that action representations are significantly more sensitive to distribution shift than visual representations. Manifold curvature estimates reveal a consistent discrepancy ($\kappa_v = 3.8 \pm 0.6$ vs. $\kappa_a = 1.3 \pm 0.2$, $p < 0.001$), confirming that the two modalities inhabit incompatible manifolds. See Appendix~\ref{app:manifold} for details.

We provide a formal analysis of why a manifold-agnostic optimization objective systematically loosens the generalization bound for the action modality. Following~\cite{sarkar2025learning}, for $L$-Lipschitz functions ($\pi_\theta$ here) on a compact Riemannian manifold with sectional curvature $\kappa$, the generalization error admits a curvature-dependent penalty $\psi(\kappa, L) = \sqrt{|\kappa|}/L$, which tightens only when the objective is geometrically matched to the target manifold. In existing unified WAMs, a single velocity-based objective calibrated to the high-curvature visual manifold $\mathcal{M}_v$ is imposed on action modality. Since $\pi_\theta$ is jointly optimized over both manifolds, its Lipschitz regularity is governed by the more complex geometry of $\mathcal{M}_v$, effectively inflating the curvature experienced by the action branch from $\kappa_a$ to $\kappa_v$. Since $\psi$ is monotonically increasing in $|\kappa|$, this forces the action-side penalty to $\psi(\kappa_v, L) \gg \psi(\kappa_a, L)$, yielding a strictly looser bound on $\mathcal{M}_a$ than a $\kappa_a$-matched objective would achieve. Under distribution shift, this looseness directly manifests as degraded action robustness, formally motivating our manifold-aware multi-objective formulation.
\section{Method}

\begin{figure}[t]
    \centering
    \includegraphics[width=0.99\linewidth]{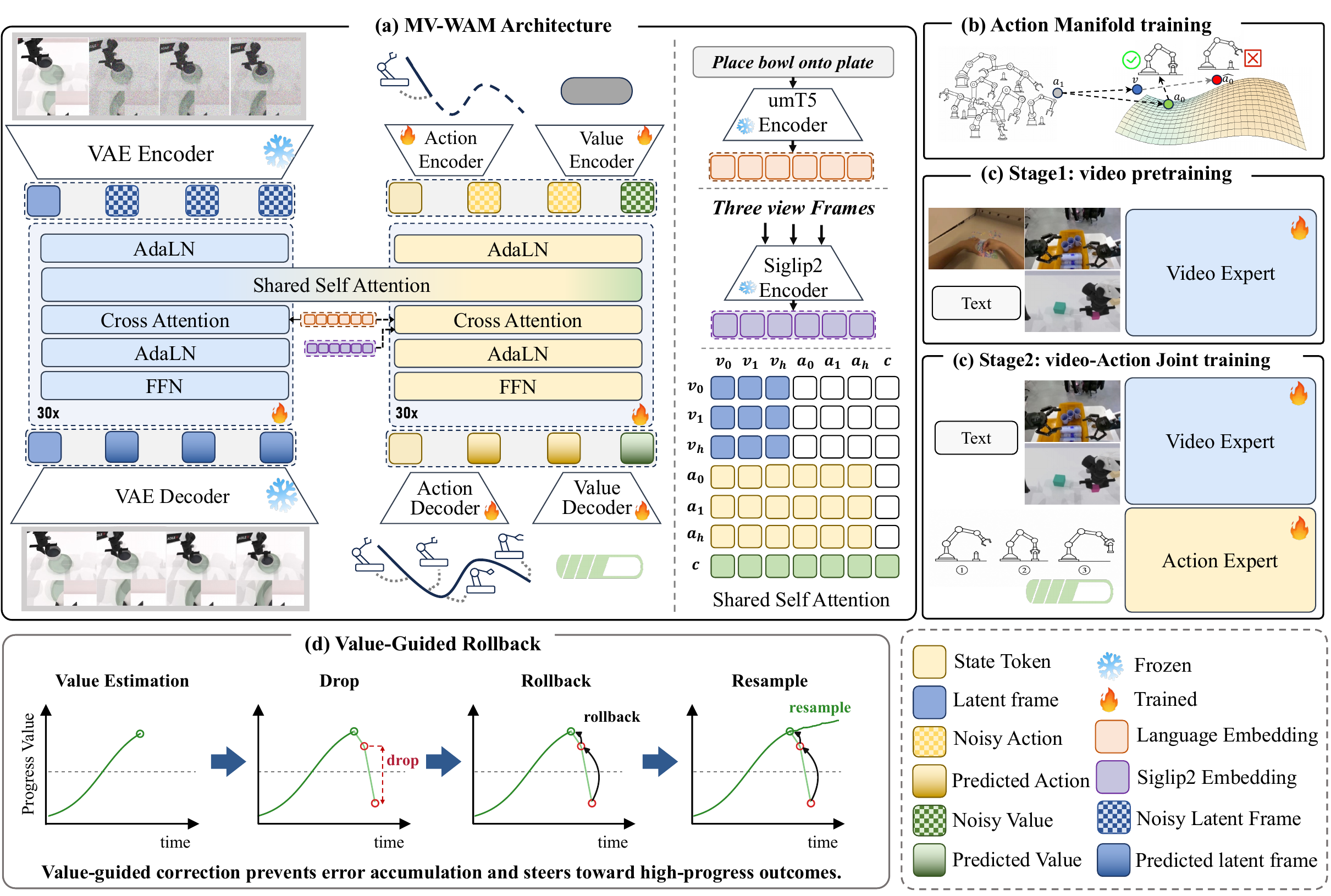}
    \caption{
        Overview of \textbf{MV-WAM}.  (a) Detailed architecture comprises a Video Expert and an Action-Value Expert. (b) Manifold-aware Training. (c) Two-stage training, video-only pretraining followed by joint video-action training. (d) Value-guided rollback for online execution correction.
    }
    \label{fig:framework}
    \vspace{-0.5cm}
\end{figure}

\subsection{Novel End-to-End Framework}
\label{sec:architecture}
To close the generalization gap arising from the structural mismatch between visual and action modalities, we propose a Manifold-Aware Unified Framework named \textbf{MV-WAM}, which unifies video world modeling, robotic action prediction, and value estimation within a single Transformer-based diffusion backbone. As illustrated in Fig.~\ref{fig:framework}, our model is built upon stacked Diffusion Transformer (DiT) blocks comprising a Video Expert and an Action-Value Expert, coupled through a Mixture-of-Transformers (MoTs)~\cite{liang2024mixture_of_transformers} mechanism that enables cross-modal interaction. Rather than adopting the plain MoTs, we extend it into a manifold-aware multi-objective MoTs.

\paragraph{Dual Branch Expert architecture}
Both the Video Expert and Action-Value Expert adopt DiT backbones with identical depth and standard internal configurations. Task instructions $\mathcal{L}$ are uniformly encoded via the umT5~\cite{chung2023unimax} text encoder and injected into both experts through cross-attention. Specifically, the Video Expert is instantiated based on WoW-1.3B~\cite{chi2025wow}. It first employs a spatiotemporal VAE to compress raw visual observation frames into compact latent representations, which are concatenated with mask tokens along the channel dimension, patched into visual tokens, and fed into the DiT backbone for video diffusion modeling. In contrast, the Action-Value Expert takes multi-view visual states $\boldsymbol{s}_t$ encoded by SigLIP2~\cite{tschannen2025siglip2} as visual condition, with language embeddings injected into odd transformer layers and SigLIP2 visual embeddings into even layers. Within this expert, action tokens and state tokens share a unified encoder-decoder architecture for action generation, while value tokens are processed by an independent encoder-decoder for separate value estimation.

For precise temporal alignment during joint modeling, the Rotary Positional Embedding (RoPE) scaling factor of the Action-Value Expert is set to $1/4$ of that used in the Video Expert. This synchronization establishes a unified temporal grid between action tokens and video latents, enabling temporally coherent cross-modal interaction. We further adopt a structured causal attention mask to regulate information flow while preserving modality-specific priors. Video tokens are restricted to self-attention only within the visual stream to maintain high-fidelity video generation, whereas action and state tokens can attend to both video context and other action tokens. value tokens attend to visual, action, and state tokens jointly, enabling holistic assessment of task progress and visual-action alignment. This asymmetric masking ensures that action generation is always conditioned on visual predictions while video generation remains uncontaminated by action signals, and that value estimation has access to the full cross-modal context necessary for reliable task assessment.

\paragraph{Manifold-Aware Dual Prediction Objectives}
Instead of adopting a single unified diffusion objective for both modalities, we tailor two distinct manifold-aware prediction losses according to their intrinsic geometric properties. Visual and action data reside on Riemannian manifolds with significantly different curvatures, and optimizing both with one shared objective leads to severe geometric mismatch and suboptimal generalization. Our framework adopts separate flow-matching transformation functions for each branch. The Video Expert, defined on the high-curvature visual manifold, directly predicts the velocity field following standard flow matching:
\begin{equation}
    \mathcal{L}_v = \mathbb{E}_{\tau,\boldsymbol{z}_0,\boldsymbol{z}_1}
    \left\| \pi_\theta^v(\boldsymbol{z}_\tau, \tau, \boldsymbol{c})
    - (\boldsymbol{z}_1 - \boldsymbol{z}_0) \right\|_2^2,
\end{equation}
where $\boldsymbol{z}_0$ denotes clean video latents, $\boldsymbol{z}_1$ is the noise prior, and $\boldsymbol{z}_\tau = (1-\tau)\boldsymbol{z}_0 + t\boldsymbol{z}_1$.
$\pi_\theta^v$ and $\pi_\theta^a$ denote the Video Expert and Action-Value Expert respectively.
For the Action-Value Expert, both action and value tokens reside on low-curvature manifolds with similar geometric structure. Using action as a representative example, the optimization objective is equivalent to directly regressing the clean action latent $\boldsymbol{a}_0$:
\begin{equation}
    \mathcal{L}_a = \mathbb{E}_{\tau,\boldsymbol{a}_0,\boldsymbol{a}_1}
    \left\| \pi_\theta^a(\boldsymbol{a}_\tau, \tau, \boldsymbol{c}) - \boldsymbol{a}_0 \right\|_2^2.
\end{equation}
The entire framework is jointly optimized by $\mathcal{L} = \lambda_v \mathcal{L}_v + \lambda_a \mathcal{L}_a$, where all balancing weights $\lambda_v$ and $\lambda_a$ are empirically set to 1. During inference, the Video Expert denoises latents from the predicted velocity field, while the Action-Value Expert directly outputs the optimized action representation. Benefiting from the manifold-aware dual objectives, MV-WAM simultaneously achieves high-fidelity visual world modeling and robust robotic action generation.

\subsection{Progress-Valued Regulation}
\label{subsec:progress_regulation}

While manifold-aware objectives ensure geometric alignment during training, they provide no explicit signal to assess task progress or detect visual-action misalignment at inference time. In the absence of such feedback, execution deviations remain undetected and unrecoverable, directly undermining the policy's generalization under distribution shift. To this end, we introduce a \textbf{Progress-Valued Regulation} mechanism that equips the Action-Value Expert with a dedicated value prediction capability, enabling the model to bias denoising trajectories toward high-progress outcomes.

\paragraph{Monte Carlo Value Estimation.}
Inspired by the paradigm of Cosmos Policy~\cite{cosmospolicy}, we estimate the task value function $c$ with Monte Carlo (MC) returns computed from executed trajectories. For each state-action pair in a trajectory, the value target is defined as the empirical discounted return accumulated from the current step $t$ to the end of the trajectory:
\begin{equation}
    \hat{c}(s_t, a_t) = \sum_{i=0}^{H-t} \gamma^i r(s_{t+i}, a_{t+i}),
\end{equation}
where $H$ is the trajectory horizon and $\gamma$ is the discount factor(we set $\gamma=0.99$). This MC target converts sparse task outcomes into a dense progress signal along the demonstration trajectory, serving as the supervision target for the value token.

\paragraph{Value-Guided Rollback for Online Execution.} 
During closed-loop execution, the predicted value token serves as a dynamic progress monitor to detect and correct execution deviations in real-time. Specifically, we maintain a rolling buffer of recent state-Action-Value triplets and track the peak value $c_{\text{peak}} = \max_{i < t} \hat{c}_i$ achieved so far. At each control step, if the newly predicted value drops significantly below this peak (i.e., $\hat{c}_t < c_{\text{peak}} - \delta$, where $\delta$ is a task-aware threshold), the system flags the current action chunk as potentially erroneous. Upon triggering, the policy initiates a \textbf{rollback-and-resample} protocol: it reverts the execution state to the cached checkpoint corresponding to $c_{\text{peak}}$, effectively discarding the faulty action sequence. From this restored high-value state, the diffusion process is re-initialized to conditionally resample a new action chunk. Leveraging the inherent stochasticity of diffusion sampling, this correction can be performed efficiently without regenerating the entire trajectory or requiring additional environment queries. This value-triggered backtracking mechanism acts as a self-correcting loop that prevents error accumulation during long-horizon tasks, ensuring that the model consistently recovers from execution deviations and navigates toward high-progress regions of the action manifold.
\section{Experiments}
\label{sec:05_experiment}

Training configuration is detailed in Section~\ref{sec:05_experiment:1_setup}. We then benchmark MV-WAM against VLA and WAM baselines on RoboTwin~2.0 in Section~\ref{sec:05_experiment:2_simulation}, including zero-shot and few-shot generalization evaluations. Ablation studies in Section~\ref{sec:05_experiment:3_ablation} validate our key design choices, followed by real-world evaluations in Section~\ref{sec:05_experiment:4_realworld}.

\subsection{Training Configuration}
\label{sec:05_experiment:1_setup}

Excluding the umT5 encoder and SigLIP2 encoder, the complete MV-WAM architecture contains 1.9B parameters. We construct a large-scale pre-training corpus by integrating diverse public robotic manipulation datasets covering a broad range of real-world scenarios, supplemented by egocentric data to strengthen the model's visual understanding and spatial generalization. Detailed dataset statistics are provided in Appendix~\ref{appendix:dataset}. We perform a two-stage training procedure: we first pre-train the Video Expert on large-scale visual data, and then fine-tune the full model on paired demonstrations.

\textbf{Video Expert pre-training.} The Video Expert comprises 30 Transformer blocks and is initialized from WoW-1.3B~\cite{chi2025wow}, pre-trained for 20k steps with a batch size of 1024, a learning rate of $1 \times 10^{-4}$, and a weight decay of $1 \times 10^{-2}$ using AdamW. Input videos are resized to $240 \times 320$ pixels.

\textbf{Joint vision-action fine-tuning.} The full model, comprising the pre-trained Video Expert and a 30-block Action-Value Expert trained from scratch, is fine-tuned for 15k steps with a batch size of 512, a learning rate of $1 \times 10^{-4}$, and an action chunk size of 32. To reduce parameter count and computational cost, the Action-Value Expert operates at a lower hidden dimension of 768 and is projected to the 1536-dimensional video-token space only inside shared self attention. We adopt a learning rate schedule that remains constant for the first 20\% of steps and decays linearly to zero thereafter. The model achieves an inference speed of \textbf{17.9} actions per second on a single A800 GPU.

\begin{table}[t]
    \caption{\textbf{Results on the RoboTwin 2.0 benchmark (See Appendix~\ref{appendix:exp_robotwin_total} for details).}
    We evaluate models under both \textit{Clean} and \textit{Random} settings. Results report task-level manipulation success rates (\%).
    $^\dagger$ denotes reproduced using the official codebase. Mean SR denotes the average success rate.
    }
    \centering
    \small
    \resizebox{\textwidth}{!}{
    \begin{tabular}{c|cc|cc|cc|cc|c|cc|cc}
    \toprule
    \multirow{3}{*}{\textbf{Models}}
    & \multicolumn{2}{c|}{\textbf{Adjust}}
    & \multicolumn{2}{c|}{\textbf{Beat Block}}
    & \multicolumn{2}{c|}{\textbf{Blocks}}
    & \multicolumn{2}{c|}{\textbf{Blocks}}
    & \multirow{3}{*}{$\cdots$}
    & \multicolumn{2}{c|}{\textbf{Turn}}
    & \multicolumn{2}{c}{\multirow{2}{*}{\textbf{Mean SR}}} \\
    & \multicolumn{2}{c|}{\textbf{Bottle}}
    & \multicolumn{2}{c|}{\textbf{Hammer}}
    & \multicolumn{2}{c|}{\textbf{Ranking RGB}}
    & \multicolumn{2}{c|}{\textbf{Ranking Size}}
    &
    & \multicolumn{2}{c|}{\textbf{Switch}}
    & \multicolumn{2}{c}{} \\
    \cmidrule(lr){2-3}
    \cmidrule(lr){4-5}
    \cmidrule(lr){6-7}
    \cmidrule(lr){8-9}
    \cmidrule(lr){11-12}
    \cmidrule(lr){13-14}
    & Clean & Rand.
    & Clean & Rand.
    & Clean & Rand.
    & Clean & Rand.
    &
    & Clean & Rand.
    & Clean & Rand. \\
    \midrule
    $\pi_0$~\cite{2024_10_31_pi0} & 90 & 56 & 43 & 21 & 19 & 5 & 7 & 1 & $\cdots$ & 27 & 23 & 46.4 & 16.3 \\
    RDT~\cite{liu2024rdt} & 81 & \textbf{75} & 77 & 37 & 3 & 0 & 0 & 0 & $\cdots$ & 35 & 15 & 34.5 & 13.7 \\
    UP-VLA~\cite{zhang2025upvla} & \textbf{100} & 17 & 66 & 16 & 38 & 0 & 21 & 0 & $\cdots$ & 43 & 26 & 52.9 & 15.2 \\
    DP~\cite{chi2025diffusion} & 97 & 0 & 42 & 0 & 0 & 0 & 1 & 0 & $\cdots$ & 36 & 1 & 28.0 & 0.6 \\
    HALO~\cite{shou2026halo} & 97 & 9 & \textbf{96} & 11 & 94 & 7 & 58 & 2 & $\cdots$ & \textbf{65} & 27 & 80.5 & 26.4 \\
    BagelVLA~\cite{hu2026bagelvla} & \textbf{100} & 14 & 87 & 16 & 84 & 4 & 45 & 2 & $\cdots$ & 49 & 30 & 75.3 & 20.5 \\
    Fast-WAM$^\dagger$~\cite{yuan2026fastwam} & 95 & 0 & 74 & 2 & 0 & 1 & 38 & 0 & $\cdots$ & 56 & 17 & 71.9 & 6.3 \\
    \midrule
    \rowcolor[HTML]{FFF0F5}
    \textbf{Ours} & 83 & 65 & 75 & \textbf{53} & \textbf{99} & \textbf{88} & \textbf{63} & \textbf{43} & $\cdots$ & 62 & \textbf{65} & \textbf{84.0} & \textbf{55.7} \\
    \bottomrule
    \end{tabular}}
    \label{tab:exp_robotwin_preview}
    \vspace{-0.5cm}
    \end{table}

\subsection{Simulation Experiments}
\label{sec:05_experiment:2_simulation}

\paragraph{Benchmark and Protocol.} We evaluate our method on the RoboTwin~2.0~\cite{chen2025robotwin} benchmark using the Aloha AgileX dual-arm embodiment. For VLA baselines, we follow the standard single-stage fine-tuning protocol, training each model on 50 \textit{Clean} expert demonstrations per task. For WAM baselines and MV-WAM, we follow the two-stage training protocol of BagelVLA~\cite{hu2026bagelvla}: the Video Expert is first pre-trained on a mixture of 50 \textit{Clean} and 500 \textit{Random} demonstrations per task, and the full model is then fine-tuned exclusively on \textit{Clean} paired demonstrations with video weights initialized from the first stage. This decoupled design deliberately introduces a distribution shift between fine-tuning and evaluation, providing a controlled measure of each model's generalization capability. All models are evaluated on 100 rollouts per task under both \textit{Clean} and \textit{Random} conditions.

\paragraph{Baselines.} 
We compare MV-WAM against seven baselines spanning two categories. For VLA models, we evaluate against DP~\cite{chi2025diffusion}, $\pi_0$~\cite{2024_10_31_pi0}, and RDT~\cite{liu2024rdt}. For WAMs, we benchmark against UP-VLA~\cite{zhang2025upvla}, BagelVLA~\cite{hu2026bagelvla}, HALO~\cite{shou2026halo}, and Fast-WAM~\cite{yuan2026fastwam}.

\paragraph{Results.}
As shown in Table~\ref{tab:exp_robotwin_preview}, MV-WAM achieves state-of-the-art performance across all tasks under both evaluation conditions, attaining average success rates of \textbf{84.0\%} and \textbf{55.7\%} on \textit{Clean} and \textit{Random}, respectively. Among VLA baselines, $\pi_0$ performs strongest at 46.4\% / 16.3\%, suggesting that standard VLA methods struggle to generalize to unseen visual perturbations when visual and action representations are learned independently.

Comparing against WAM baselines further validates the importance of modality-aware cross-modal alignment. 
While HALO, BagelVLA, and Fast-WAM achieve strong \textit{Clean} performance (80.5\%, 75.3\%, and 71.9\%), their \textit{Random} performance remains substantially lower (26.4\%, 20.5\%, and 6.3\%), indicating that visual prediction alone does not sufficiently transfer to action-level robustness. 
MV-WAM surpasses the strongest WAM baseline, HALO, by 3.5\% on \textit{Clean} and 29.3\% on \textit{Random}, and outperforms Fast-WAM by 12.1\% on \textit{Clean} and 49.4\% on \textit{Random}. 
Moreover, MV-WAM shows a notably smaller relative drop from \textit{Clean} to \textit{Random} (33.7\%) compared to HALO (67.2\%), BagelVLA (72.8\%), and Fast-WAM (91.2\%), confirming that geometry-aware prediction targets yield substantially more robust representations under distribution shift.

\begin{table}[H]
\caption{\textbf{Zero- and few-shot generalization on 10 unseen tasks.}
We evaluate MV-WAM under both \textit{Clean} and \textit{Random} settings across 0-shot, 1-shot,
5-shot, and 10-shot conditions. Results report task-level manipulation success rates (\%). Mean SR denotes the average success rate across 10 tasks.
}
\centering
\footnotesize
\setlength{\tabcolsep}{4pt}
\begin{tabular}{c|cc|cc|cc|cc}
\toprule
\multirow[c]{2}{*}{\textbf{Task}}
& \multicolumn{2}{c|}{\textbf{0-shot}}
& \multicolumn{2}{c|}{\textbf{1-shot}}
& \multicolumn{2}{c|}{\textbf{5-shot}}
& \multicolumn{2}{c}{\textbf{10-shot}} \\
\cmidrule(lr){2-3}
\cmidrule(lr){4-5}
\cmidrule(lr){6-7}
\cmidrule(lr){8-9}
& Clean & Rand.
& Clean & Rand.
& Clean & Rand.
& Clean & Rand. \\
\midrule
Blocks Ranking Size   & 48 &  50 & 52 &  62 & 56 &  77 & 63 &  69 \\
Click Bell            & 96 &  93 & 99 & 100 & 98 &  99 & 100 & 99 \\
Place Dual Shoes      &  0 &   0 & 26 &  20 & 22 &  36 & 22 &  38 \\
Move Stapler Pad      & 26 &  25 & 21 &  25 & 47 &  38 & 30 &  30 \\
Shake Bottle Horiz.   & 98 & 100 & 97 & 100 & 95 & 100 & 100 & 100 \\
Pick Dual Bottles     & 60 &  82 & 14 &   9 & 89 &  74 & 86 &  70 \\
Place A2B Right       & 68 &  61 & 86 &  70 & 87 &  79 & 91 &  83 \\
Place Bread Skillet   & 30 &  13 & 28 &  26 & 82 &  67 & 81 &  73 \\
Place Cans Plasticbox & 50 &  68 & 72 &  77 & 92 &  95 & 98 &  96 \\
Stack Blocks Three    & 80 &  48 & 78 &  74 & 94 &  92 & 100 &  88 \\
\midrule
\rowcolor[HTML]{FFF0F5}
\textbf{Mean SR}
& 55.6 & 54.0
& 57.3 & 56.3
& 76.2 & \textbf{75.7}
& \textbf{77.1} & 74.6 \\
\bottomrule
\end{tabular}
\vspace{-10pt}
\label{tab:zeroshot}
\end{table}

\paragraph{Zero-Shot and Few-Shot Generalization.}
To evaluate the transferability of MV-WAM to unseen tasks, we conduct zero-shot and few-shot Out-of-Distribution (OOD) experiments on RoboTwin~2.0. MV-WAM is trained on 40 tasks (a mixture of 50 Clean and 500 Random demonstrations per task, totally 22000 demonstrations) and evaluated on 10 OOD tasks. In the zero-shot setting, no demonstrations from test tasks are provided during fine-tuning. In the few-shot setting, we supply 1, 5, or 10 demonstrations per unseen task and fine-tune accordingly.


As shown in Table~\ref{tab:zeroshot}, MV-WAM already achieves strong performance in the zero-shot setting (55.6\% / 54.0\% on \textit{Clean} / \textit{Random}), demonstrating generalization to novel tasks without any task-specific supervision. Performance improves substantially at 10-shot, reaching 77.1\% on \textit{Clean} and 74.6\% on \textit{Random}, confirming that MV-WAM provides a strong foundation for data-efficient generalization to unseen manipulation tasks.

\subsection{Ablation Study}
\label{sec:05_experiment:3_ablation}



\begin{figure}[t]
    \centering
    \includegraphics[width=1.00\linewidth]{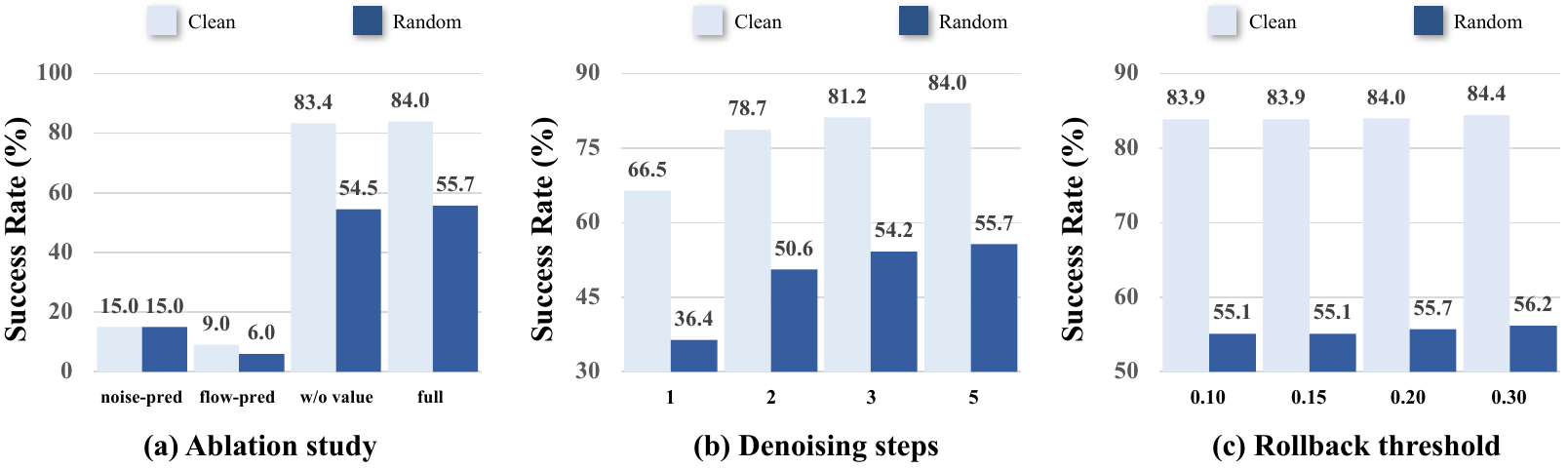}
    \caption{
        \textbf{Ablation study results.}
        (a) Manifold-aware prediction targets.
        (b) Effect of denoising steps.
        (c) Sensitivity to rollback threshold.
    }
    \label{fig:ablation}
\end{figure}

\paragraph{Manifold-Aware Prediction Targets.} All configurations use the same training setting and are trained for 15k steps, with only the prediction target assignment changed. As shown in Figure~\ref{fig:ablation}(a), the optimal configuration assigns direct $x_0$-prediction to the action branch and velocity prediction to the video branch, achieving 84.0\% / 55.7\% on Clean / Random. Both noise-pred and flow-pred baselines collapse substantially under Random conditions (15.0\% and 6.0\%), confirming that applying a single unified objective across modalities severely degrades action robustness under distribution shift. These results validate our core design premise: robot actions reside on a low-dimensional, smooth manifold where direct state regression provides a more stable training signal, while the higher-curvature visual manifold favors velocity-based prediction.


\paragraph{Denoising Steps.} We analyze the effect of the number of shared denoising steps applied to both the Video Expert and Action-Value Expert on task performance. As shown in Figure~\ref{fig:ablation}(b), performance improves consistently with additional steps but the gains diminish rapidly: MV-WAM with 3 steps already recovers over 96\% of the Clean performance attained at 5 steps, while reducing denoising cost by 40\%. Even with a single step, MV-WAM maintains competitive success rates (66.5\% / 36.4\%), demonstrating that the jointly trained representations are robust to coarse denoising schedules. We adopt 5 denoising steps as our default throughout all experiments.

\paragraph{Progress-Value Tokens.} We further isolate the contribution of the progress-value prediction branch and the sensitivity of the inference-time rollback threshold~$\delta$. As shown in Figure~\ref{fig:ablation}(c), value tokens consistently improve over the no-value baseline (83.4\% / 54.5\%) across all tested thresholds, confirming that the task-completion signal provides complementary supervisory information for policy learning. Performance varies only marginally across $\delta$ values; we adopt $\delta{=}0.20$ as our default (84.0\% / 55.7\%), indicating that the rollback mechanism is robust to this hyperparameter.

\subsection{Real-World Experiments}
\label{sec:05_experiment:4_realworld}

\paragraph{Data Collection.} To evaluate the practical applicability of MV-WAM, we deploy our framework on a TienKung dual-arm robot platform across four daily manipulation tasks of increasing difficulty: \textit{Pick Backbag \& Coffee}, \textit{Drop Cloth}, \textit{Pick Cloth}, and \textit{Fold Cloth}, spanning rigid object grasping, cloth handling, and fine-grained deformable manipulation. We curate a dataset of 100 demonstrations per task, collected via human teleoperation at 30 FPS. Details of the hardware configuration and camera setup are provided in Appendix~\ref{appendix:realworld_setup}.

\paragraph{Training and Evaluation Protocol.} We train MV-WAM following the two-stage protocol described in Section~\ref{sec:05_experiment:1_setup}, with three-view visual inputs from cameras mounted at the wrist and third-person perspectives of the TienKung platform. We compare against two representative baselines: (i)~$\pi_0$~\cite{2024_10_31_pi0}, a state-of-the-art generalist VLA model, and (ii)~RDT~\cite{liu2024rdt}, a large-scale diffusion-based robot foundation model. For fair comparison, all methods are initialized from their official pretrained checkpoints and fine-tuned with identical demonstration sets under full fine-tuning, then evaluated over 10 trials per task under varying object configurations.

\begin{figure}
    \centering
    \includegraphics[width=0.85\linewidth]{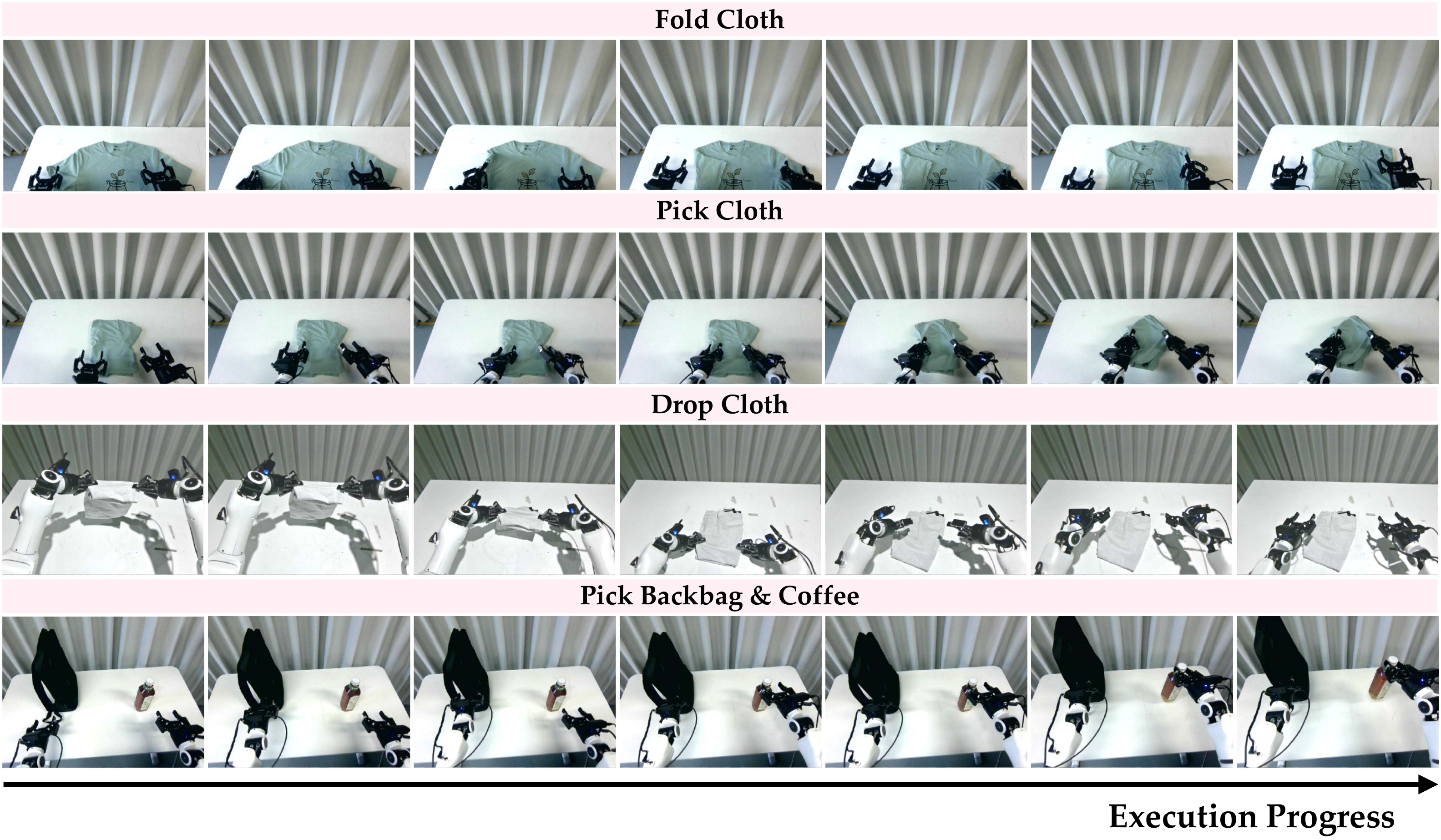}
    \caption{
        \textbf{Real-World Execution Visualization.}
        Execution progress across all four real-world manipulation tasks.
    }
    \label{fig:real_world_execution_progress}
\end{figure}

\begin{table}[t]
\caption{\textbf{Real-world experiment results on a TienKung robot.}
We report success rates (\%) over 10 trials per task.
}
\centering
\footnotesize
\begin{tabular}{l|cccc|c}
\toprule
\textbf{Method} & \shortstack{Pick Backbag} & \shortstack{Drop Cloth} & \shortstack{Pick Cloth} & \shortstack{Fold Cloth} & \textbf{Mean SR} \\
\midrule
$\pi_0$~\cite{2024_10_31_pi0} & 50 & 60 & 50 & 10 & 42.5 \\
RDT~\cite{liu2024rdt}         & 30 & 40 & 60 & 0  & 32.5 \\
\midrule
\rowcolor[HTML]{FFF0F5}
\textbf{MV-WAM (Ours)} & \textbf{90} & \textbf{100} & \textbf{100} & \textbf{20} & \textbf{77.5} \\
\bottomrule
\end{tabular}
\label{tab:realworld}
\end{table}

\paragraph{Results.} As shown in Table~\ref{tab:realworld} and Figure~\ref{fig:real_world_execution_progress}, MV-WAM achieves the highest success rates across all four tasks, attaining a mean SR of \textbf{77.5\%} and outperforming $\pi_0$ (42.5\%) and RDT (32.5\%) by substantial margins. On structured tasks such as \textit{Drop Cloth} and \textit{Pick Cloth}, MV-WAM achieves a perfect score of 100\%, while both baselines fall below 70\%, demonstrating that jointly trained visual-action representations yield more reliable rigid and semi-rigid manipulation. The gap is most pronounced on \textit{Fold Cloth}, the most demanding task requiring precise spatial reasoning and deformable object manipulation: MV-WAM achieves 20\% compared to 10\% for $\pi_0$ and 0\% for RDT, confirming that geometry-aware cross-modal alignment is especially critical for fine-grained contact-rich tasks. Overall, these results demonstrate that the advantages of MV-WAM observed in simulation transfer effectively to physical deployment, and that grounding action generation in a jointly trained world model produces more robust manipulation policies across diverse real-world scenarios.

\section{Conclusion}

We presented MV-WAM, a unified world action model that jointly generates visual predictions, robot actions, and value estimates within a single MoTs-based diffusion backbone, while treating visual and action modalities as geometrically distinct through manifold-aware objectives and cross-modality causal masking. Experiments on RoboTwin 2.0 and real-robot platforms demonstrate that MV-WAM achieves state-of-the-art performance under both in-domain and OOD settings across simulation and physical deployment. These results suggest that closing the generalization gap requires not deeper integration, but more principled geometric alignment between modalities.

\paragraph{Limitations} First, MV-WAM has not been validated at larger model scales, leaving the scalability of the proposed mechanisms an open question. Second, the Monte Carlo Value estimates may be noisy under sparse rewards, potentially limiting rollback reliability in long-horizon tasks.

\bibliographystyle{plain}
\bibliography{neurips_2026}
\clearpage

\appendix


\section{Related Work}

\paragraph{Vision-Language-Action Models.} Early imitation learning approaches such as ACT~\cite{zhao2023learning_act} and Diffusion Policy~\cite{chi2025diffusion} demonstrated the effectiveness of expressive policy architectures for manipulation, but rely on task-specific designs that generalize poorly across diverse scenarios and unseen visual conditions. VLA models address this by leveraging pretrained vision-language models~\cite{radford2021learning, alayrac2022flamingo} to parameterize manipulation policies, enabling natural language instruction following and broader semantic generalization. Several pioneering works~\cite{brohan2022rt, driess2023palm, 2024_9_05_OpenVLA} demonstrate that training on large-scale and diverse robot demonstration datasets can substantially enhance the cross-task generalization of VLA models. To capture more complex action distributions, subsequent works augment VLA policies with continuous generative heads, encompassing diffusion-based denoising formulations~\cite{wen2024diffusion, 2025_3_13_HybridVLA} and flow-matching alternatives that enable more efficient continuous action synthesis~\cite{2024_10_31_pi0, 2025_4_22_pi0_5}. Despite these advances, VLA models remain brittle under environmental perturbations such as novel lighting, background textures, and unseen object configurations, as their visual representations are learned from demonstration corpora that cannot exhaustively cover real-world deployment diversity. MV-WAM addresses this limitation by incorporating a large-scale video-generative prior that broadens visual coverage well beyond what labeled demonstrations can provide.


\paragraph{World Action Models.} A growing line of work seeks to improve policy generalization by incorporating world modeling into the action learning loop. UP-VLA~\cite{zhang2025upvla} and Video Prediction Policy~(VPP)~\cite{hu2024video_prediction_policy} introduce visual prediction as an auxiliary objective to enrich visual representations, while BagelVLA~\cite{hu2026bagelvla} interleaves linguistic planning and visual forecasting sequentially before action generation. Fast-WAM~\cite{yuan2026fastwam} demonstrates that the primary benefit of world modeling lies in representation learning during training rather than explicit future imagination at inference time. More recent work such as Motus~\cite{bi2025motus} and HALO~\cite{shou2026halo} adopt Mixture-of-Transformers architectures~\cite{liang2024mixture_of_transformers} to unify understanding, visual prediction, and action generation within a single model, achieving strong performance on challenging manipulation benchmarks. 

\section{Pre-training Dataset Details}
\label{appendix:dataset}

We summarize the datasets used in our two-stage pre-training pipeline in Table~\ref{tab:dataset_hours}. The corpus covers a broad range of real-world manipulation scenarios collected through both human demonstration and robot teleoperation, totaling approximately 7,500 hours of interaction data.

\begin{table}[H]
\caption{\textbf{Datasets used in embodied data pre-training.}
We list each dataset and its estimated data collection duration in hours.
}
\centering
\small
\begin{tabular}{lc}
\toprule
\textbf{Dataset} & \textbf{Hours} \\
\midrule
EgoDex~\cite{hoque2025egodex}       & 800   \\
Agibot~\cite{bu2025agibot}          & 2,500 \\
RoboMind~\cite{wu2024robomind}       & 1,000 \\
RoboCoin~\cite{wu2025robocoin}       & 1,000 \\
Open X-Embodiment~\cite{o2024open}  & 2,000 \\
Internal private data               & 200   \\
\midrule
\textbf{Total}                      & \textbf{7,500} \\
\bottomrule
\end{tabular}
\label{tab:dataset_hours}
\end{table}

\section{Real-World Experiment Setup}
\label{appendix:realworld_setup}

\subsection{TienKung Dual-Arm Robot Setup}

\begin{figure}[H]
    \centering
    \includegraphics[width=0.80\linewidth]{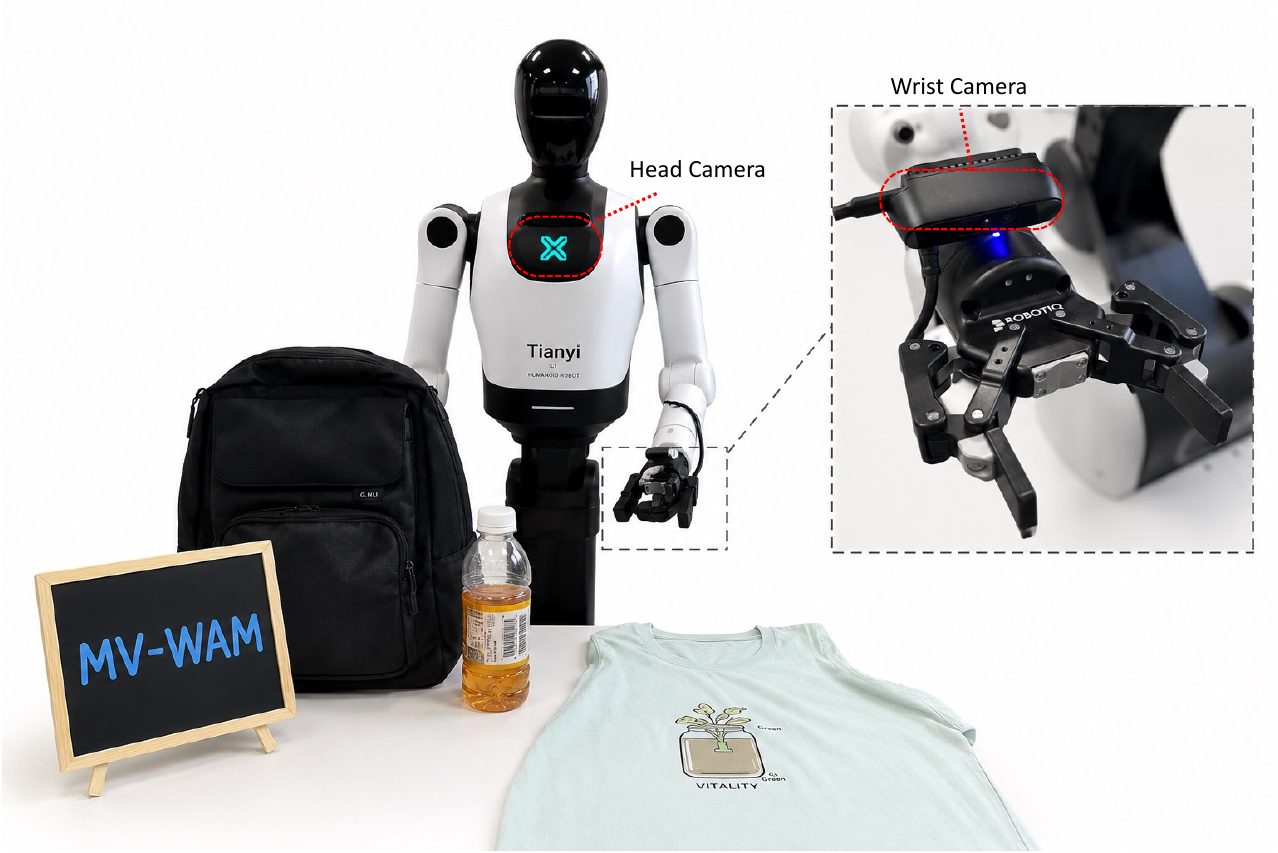}
    \caption{
        \textbf{Real-world robot experiment setup.}
        The TienKung dual-arm platform with two wrist-mounted cameras and one stationary head camera for three-view RGB observation.
    }
    \label{fig:realworld_setup}
\end{figure}

As shown in Figure~\ref{fig:realworld_setup}, the physical deployment of MV-WAM utilizes a TienKung dual-arm robot platform designed for dexterous bimanual manipulation. The perceptual system consists of three Orbbec Gemini 330 RGB-D cameras: two wrist-mounted cameras providing close-range views of each end-effector for fine-grained contact reasoning, and one stationary third-person camera offering a global workspace perspective for spatial planning and task progress monitoring.

All demonstration trajectories were collected via human teleoperation at 30 FPS, with 300 high-quality demonstrations curated per task. We adopt a unified action space $\boldsymbol{a} \in \mathbb{R}^{16}$, where each of the two arms contributes 7-DoF joint position commands (delta angles) and 1-DoF gripper state:
\begin{equation}
    \boldsymbol{a} = [\Delta\theta^R_{1:7},\, g^R,\, \Delta\theta^L_{1:7},\, g^L]
    \in \mathbb{R}^{16}.
\end{equation}

During training, the three-view RGB observations are encoded by SigLIP2 as the visual condition input to the Action-Value Expert, following the same protocol described in Section~\ref{sec:05_experiment:1_setup}. All methods are initialized from their official pretrained checkpoints and fine-tuned on the collected demonstrations under identical settings for fair comparison.

\section{Generalization Analysis in Unified World Action Models}

\subsection{The Generalization Gap Persists Across Architectures}
\label{app:manifold}

\begin{wrapfigure}[10]{r}{0.35\textwidth}
    \vspace{-0.5em}
    \centering
    \includegraphics[width=\linewidth]{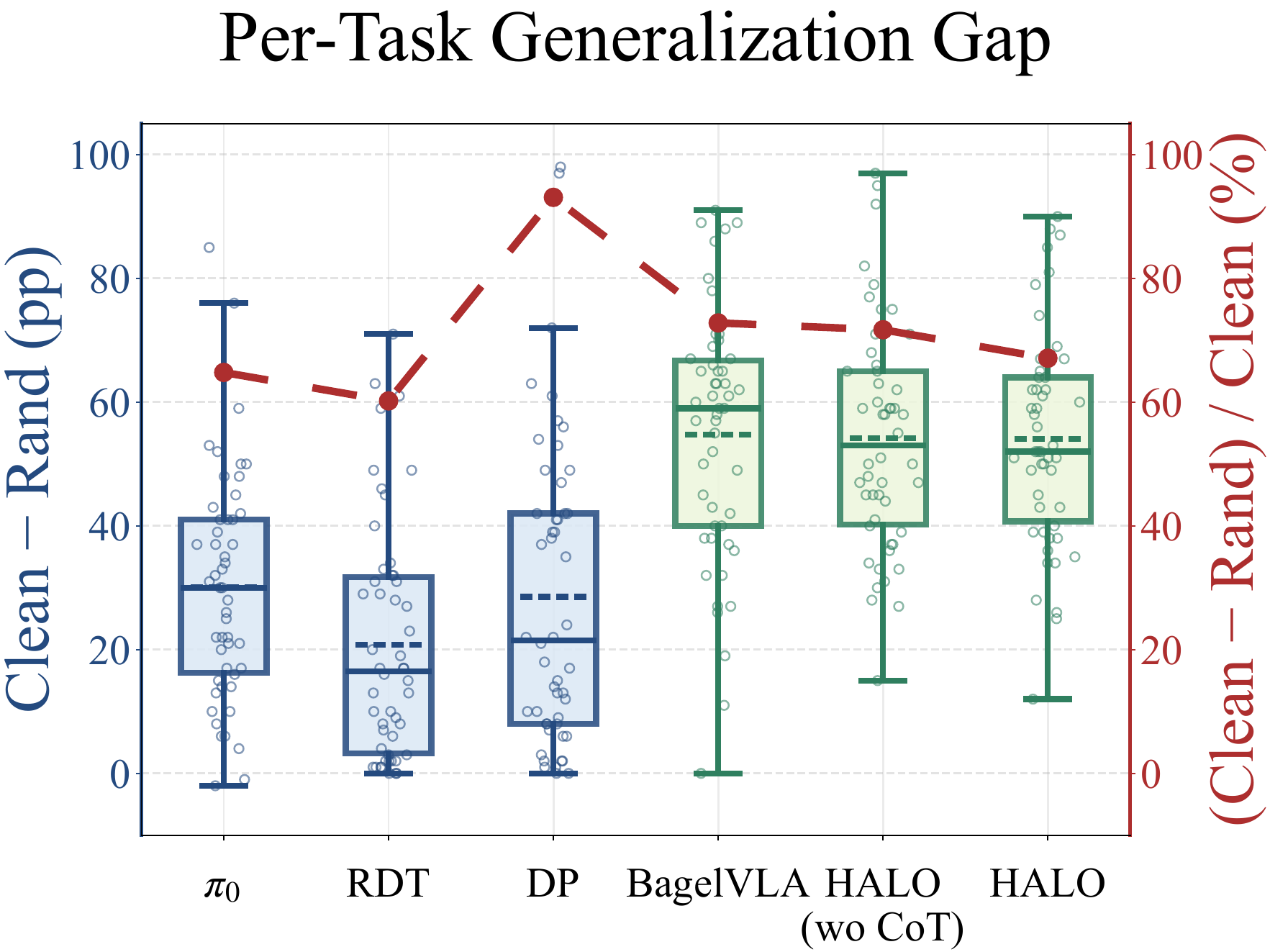}
    \caption{
        Generalization Gap for models.
    }
    \label{fig:generalization_gap}
    \vspace{-1em}
\end{wrapfigure}

The key distinction between VLAs and WAMs lies in whether visual state prediction and action generation are jointly optimized within a shared latent space. Unified models, including BagelVLA and HALO, achieve substantially higher success rates under clean conditions (75.3\%–80.5\%) compared to non-unified baselines (28.0\%–46.4\%). Yet a natural question arises: \textbf{\textit{does this shared latent space actually bridge the generalization gap?}}


Empirical evidence suggests otherwise. All models are evaluated on RoboTwin 2.0, where training demonstrations consist exclusively of clean-condition action trajectories without any domain-randomized action supervision. As shown in Fig. \ref{fig:generalization_gap}, the gap retention rate $(\text{Clean} - \text{Random}) / \text{Clean}$ remains consistently high across both unified and non-unified architectures, ranging from 60.2\% to 93.1\% for non-unified models and 67.1\% to 72.8\% for unified models. More critically, the absolute clean-to-random gap of unified models (54.0 to 54.8 percentage points) substantially exceeds that of non-unified baselines (20.8 to 30.1 percentage points), indicating that out-of-distribution robustness not only fails to improve but degrades in absolute terms. These results suggest that visual and action generalization remain largely decoupled even when optimized within a shared latent space, and that spatiotemporal priors acquired through visual prediction do not effectively transfer to action-level robustness under environmental perturbation.


\begin{wrapfigure}[14]{l}{0.47\textwidth}
    \vspace{-0.5em}
    \centering
    \includegraphics[width=\linewidth]{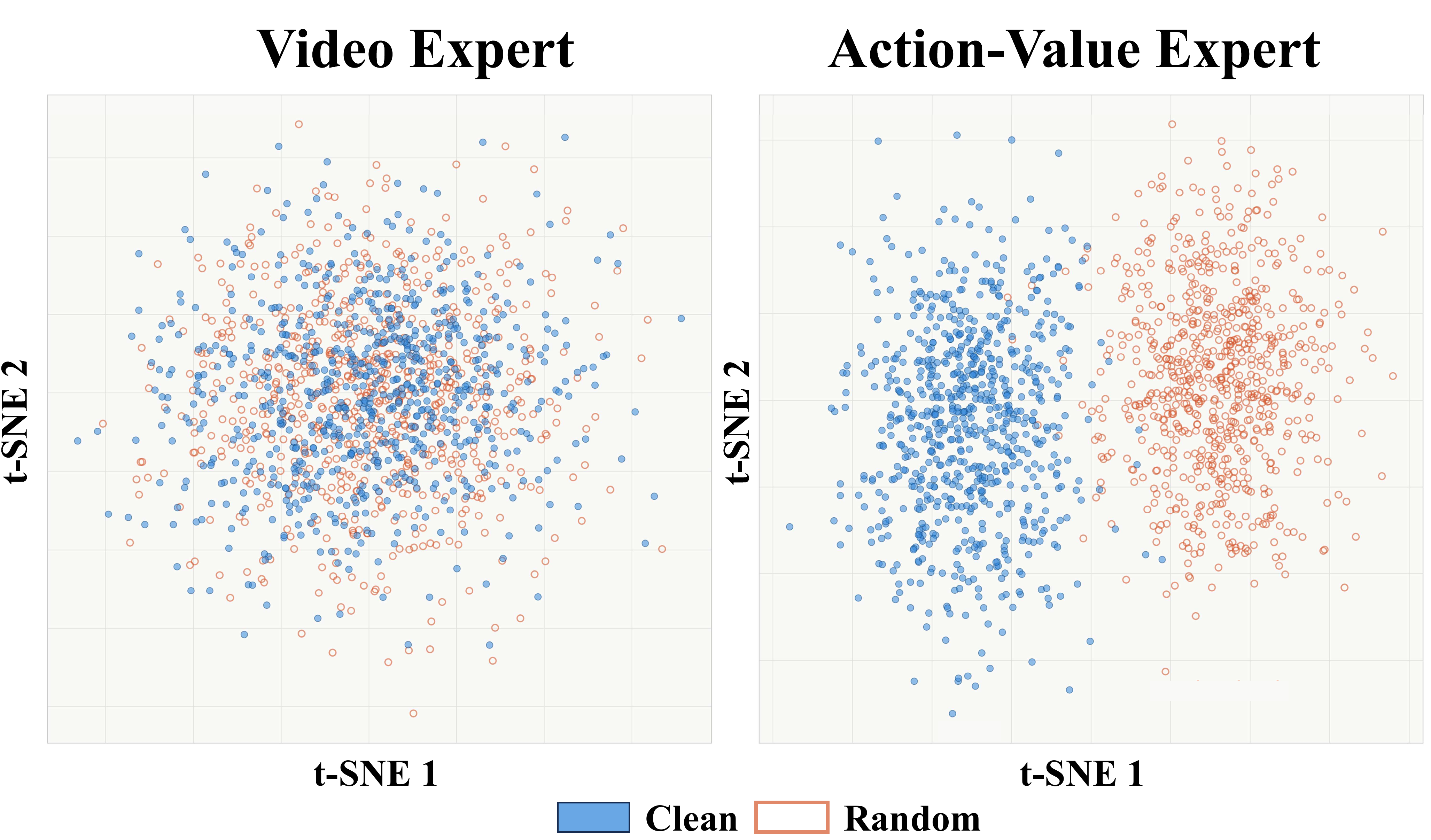}
    \caption{
        t-SNE visualization of visual and action expert representations
        after per-task mean centering.
    }
    \label{fig:tsne}
    \vspace{-1em}
\end{wrapfigure}

We probe whether this failure stems from a fundamental asymmetry in how visual and action representations respond to distributional shift. Specifically, we extract features from both the visual and action experts at the shared self-attention layer. For each of the 50 tasks, we randomly sample 3 episodes and 5 frames per episode under both clean and domain-randomized conditions, regardless of task success, yielding approximately 750 points per condition. Since the number of tokens differs between the visual and action experts, we apply mean pooling over all tokens within each frame to obtain a fixed-dimensional representation per frame. To isolate condition-level variation from inter-task variation, we apply per-task mean centering by subtracting the mean feature vector of each task before aggregating all samples for t-SNE visualization (perplexity = 50).

The t-SNE results reveal a clear representational asymmetry between the two experts as shown in Fig.~\ref{fig:tsne}. Visual expert features under clean and domain-randomized conditions remain largely intermingled, indicating that the visual representations are robust to environmental perturbation. In contrast, action expert features form two clearly separated clusters under the same perturbations, demonstrating that the action representations are significantly more sensitive to distributional shift. This asymmetry suggests that while unified modeling enables the visual latent space to generalize across conditions, this robustness does not propagate to the action latent space, confirming that visual and action generalization remain decoupled even within a shared latent space. \textit{\textbf{Joint optimization in a shared latent space is necessary but not sufficient for generalization transfer across modalities.}}

The representational asymmetry observed above naturally raises a deeper question: does this reflect an intrinsic difference in the underlying geometry of the two modalities? Manifold curvature at each sample point is estimated via the ratio of geodesic distance to Euclidean distance between each paired clean and randomized sample, where a higher ratio indicates greater local curvature. Since the visual observation space is high-dimensional, we first apply PCA to reduce each frame to 50 dimensions before constructing a 10-nearest-neighbor graph for geodesic distance approximation via shortest-path computation. For the action modality, curvature is computed directly in the native action space without dimensionality reduction, as the action vector is inherently low-dimensional. Per-task curvature estimates are obtained by averaging over all valid paired samples within each task, and statistical significance is assessed via the Mann-Whitney U test across the 50 tasks.

\begin{wraptable}{l}{0.45\textwidth}
    \vspace{-1em}
    \centering
    \caption{Manifold curvature estimates for visual and action modalities across 50 tasks.}
    \label{tab:curvature}
    \begin{tabular}{lcc}
        \toprule
        Modality & $\kappa$ (mean $\pm$ std) & $p$-value \\
        \midrule
        Visual   & 3.8 $\pm$ 0.6 & \multirow{2}{*}{$<$0.001} \\
        Action   & 1.3 $\pm$ 0.2 & \\
        \bottomrule
    \end{tabular}
    \label{tab:manifolds}
    \vspace{-0.5em}
\end{wraptable}

As shown in Table~\ref{tab:manifolds}, the visual manifold exhibits significantly higher curvature than the action manifold across all 50 tasks, and this difference is consistent regardless of task type or perturbation condition. Critically, since this measurement is performed entirely in raw data space and is independent of any model or training procedure, it confirms that \textbf{\textit{the geometric discrepancy between the two modalities is intrinsic rather than learned}}. This finding suggests that the failure of generalization transfer in unified models is not a training artifact, but a structural consequence of forcing two geometrically incompatible manifolds into a shared latent space under a single optimization objective.

\subsection{Manifold-Aware Optimization as a Necessary Condition}

Building on the curvature discrepancy established above, we now provide a formal analysis of why a manifold-agnostic optimization objective systematically loosens the generalization bound for the action modality. 
While our empirical proxy (the geodesic-to-Euclidean distance ratio) is not a direct estimate of sectional curvature, the consistent ordering $\kappa_v \gg \kappa_a$ across all 50 tasks provides the structural foundation required for this analysis. 
Following \cite{sarkar2025learning}, for a class of $L$-Lipschitz functions (here, the policy $\pi_\theta$) on a compact Riemannian manifold with sectional curvature $\kappa$, the generalization error admits a curvature-dependent penalty $\psi(\kappa, L) = \sqrt{|\kappa|}/L$. 
Crucially, this bound tightens only when the optimization objective is geometrically matched to the target manifold's intrinsic structure.

In existing unified models, a single velocity-based objective calibrated to the high-curvature visual manifold $\mathcal{M}_v$ is imposed uniformly across modalities. 
Because $\pi_\theta$ is jointly optimized over both manifolds, its overall Lipschitz regularity is governed by the more complex geometry of $\mathcal{M}_v$. 
This shared parameterization effectively inflates the curvature experienced by the action branch from its native $\kappa_a$ to $\kappa_v$. 
Since $\psi$ is monotonically increasing in $|\kappa|$, this structural mismatch forces the action-side penalty to $\psi(\kappa_v, L) \gg \psi(\kappa_a, L)$, yielding a strictly looser bound on $\mathcal{M}_a$ than a $\kappa_a$-matched objective would achieve. 
Under distributional shift, this looseness directly manifests as degraded action robustness, formally motivating our manifold-aware multi-objective formulation that independently matches each modality's optimization to its intrinsic geometry.

\section{Algorithm}
Continuous Flow Matching (CFM) defines a deterministic linear interpolation path
between target data $x_0 \sim p_{\text{data}}$ and Gaussian noise $x_1 \sim \mathcal{N}(\mathbf{0}, \mathbf{I})$:
\begin{equation}
x_t = (1-t)x_0 + t x_1, \quad t \in [0,1].
\end{equation}
The three common prediction parameterizations differ only in the supervised target and in how the inference-time velocity is reconstructed from the model output. Table~\ref{tab:fm_parameterizations} summarizes these equivalent forms.

\begin{table}[ht]
\centering
\caption{Prediction parameterizations for continuous flow matching.}
\label{tab:fm_parameterizations}
\resizebox{\textwidth}{!}{
\begin{tabular}{lccc}
\toprule
\textbf{Parameterization} & \textbf{Model output} & \textbf{Training loss} & \textbf{Inference velocity} \\
\midrule
$v$-prediction 
& $\hat{v}_\theta(x_t,t)$
& $\mathbb{E}\left[\left\|\hat{v}_\theta(x_t,t) - (x_1-x_0)\right\|_2^2\right]$
& $v_\theta(x_t,t)=\hat{v}_\theta(x_t,t)$ \\
$x_0$-prediction
& $\hat{x}_{0,\theta}(x_t,t)$
& $\mathbb{E}\left[\left\|\hat{x}_{0,\theta}(x_t,t)-x_0\right\|_2^2\right]$
& $v_\theta(x_t,t)=x_1-\hat{x}_{0,\theta}(x_t,t)$ \\
Noise-prediction
& $\hat{x}_{1,\theta}(x_t,t)$
& $\mathbb{E}\left[\left\|\hat{x}_{1,\theta}(x_t,t)-x_1\right\|_2^2\right]$
& $v_\theta(x_t,t)=\dfrac{\hat{x}_{1,\theta}(x_t,t)-x_t}{1-t}$ \\
\bottomrule
\end{tabular}}
\end{table}

In practice, the denominators in the $x_0$- and noise-prediction forms are clipped near the endpoints of the ODE trajectory for numerical stability.

\section{Discussion}

\paragraph{Relation to prior $x_0$-prediction methods.}
Although MV-WAM uses an $x_0$-style target for the action branch, this design differs from prior endpoint-prediction formulations such as pixel MeanFlow and Abot-M0. These methods predict a clean endpoint at the network-output level, but define supervision in the velocity space. Under our convention $x_t=(1-t)x_0+t x_1$, this yields
\begin{equation}
\left\| \frac{x_t-\hat{x}_{0,\theta}}{t} - \frac{x_t-x_0}{t} \right\|_2^2
= \frac{1}{t^2}\left\|\hat{x}_{0,\theta}-x_0\right\|_2^2 ,
\end{equation}
which is an endpoint regression loss reweighted by $1/t^2$. This weighting is suitable when the objective is used as a single generative loss, because it emphasizes endpoint accuracy near the data side of the flow trajectory. In MV-WAM, however, action prediction is jointly optimized with video generation. Directly adopting the velocity-space $x_0$ loss would cause the action loss to diverge as $t\rightarrow 0$, allowing the low-dimensional action branch to dominate the total objective and disrupt the intended balance with the video loss.

We therefore use a modified $x_0$-prediction objective for the action branch, supervising the clean action endpoint directly without the singular $1/t^2$ reweighting. The velocity induced by the predicted endpoint is still used for sampling, but the training signal remains bounded and comparable to the video objective. This preserves the geometric motivation of $x_0$-prediction on the locally smooth action manifold while avoiding instability in the joint video-action training regime.

The same motivation also applies to the progress-value token. Rather than treating value as a high-dimensional visual reconstruction signal, MV-WAM introduces it as another low-dimensional manifold variable aligned with the action space. The value token estimates task progress and regulates action generation, providing a compact signal for detecting low-progress action chunks and triggering value-guided rollback.

\section{Simulation Results}
\label{appendix:exp_robotwin_total}

\begin{table*}[t]
    \caption{\textbf{Results on the RoboTwin 2.0 benchmark.}
    We evaluate models under both clean and randomized settings. Results report task-level manipulation success rates (\%).
    The last row reports the average success rate across all tasks.
    }
    \centering
    \Large
    \resizebox{\textwidth}{!}{
    \begin{tabular}{c|cc|cc|cc|cc|cc|cc|cc|cc}
    \toprule
    \multirow[c]{2}{*}{\textbf{Task}}
    & \multicolumn{2}{c|}{\textbf{DP}~\cite{chi2025diffusion}}
    & \multicolumn{2}{c|}{\textbf{RDT}~\cite{liu2024rdt}}
    & \multicolumn{2}{c|}{\textbf{$\pi_0$~\cite{2024_10_31_pi0}}}
    & \multicolumn{2}{c|}{\textbf{UP-VLA}~\cite{zhang2025upvla}}
    & \multicolumn{2}{c|}{\textbf{BagelVLA}~\cite{hu2026bagelvla}}
    & \multicolumn{2}{c|}{\textbf{HALO}~\cite{shou2026halo}}
    & \multicolumn{2}{c|}{\textbf{Fast-WAM}~\cite{yuan2026fastwam}}
    & \multicolumn{2}{c}{\textbf{Ours}} \\
    \cmidrule(lr){2-3}
    \cmidrule(lr){4-5}
    \cmidrule(lr){6-7}
    \cmidrule(lr){8-9}
    \cmidrule(lr){10-11}
    \cmidrule(lr){12-13}
    \cmidrule(lr){14-15}
    \cmidrule(lr){16-17}
    & Clean & Rand.
    & Clean & Rand.
    & Clean & Rand.
    & Clean & Rand.
    & Clean & Rand.
    & Clean & Rand.
    & Clean & Rand.
    & Clean & Rand. \\
    \midrule
    Adjust Bottle & 97 & 0 & 81 & \textbf{75} & 90 & 56 & \textbf{100} & 17 & \textbf{100} & 14 & 97 & 9 & 95 & 0 & 83 & 65 \\
    Beat Block Hammer & 42 & 0 & 77 & 37 & 43 & 21 & 66 & 16 & 87 & 16 & \textbf{96} & 11 & 74 & 2 & 75 & \textbf{53} \\
    Blocks Ranking RGB & 0 & 0 & 3 & 0 & 19 & 5 & 38 & 0 & 84 & 4 & 94 & 7 & 0 & 1 & \textbf{99} & \textbf{88} \\
    Blocks Ranking Size & 1 & 0 & 0 & 0 & 7 & 1 & 21 & 0 & 45 & 2 & 58 & 2 & 38 & 0 & \textbf{63} & \textbf{43} \\
    Click Alarmclock & 61 & 5 & 61 & 12 & 63 & 11 & 69 & \textbf{41} & 85 & 20 & 83 & 14 & \textbf{100} & 38 & 90 & 24 \\
    Click Bell & 54 & 0 & 80 & 9 & 44 & 3 & 54 & \textbf{72} & \textbf{100} & 35 & \textbf{100} & 10 & \textbf{100} & 22 & 95 & 43 \\
    Dump Bin Bigbin & 49 & 0 & 64 & 32 & 83 & 24 & 81 & 35 & 91 & 51 & 93 & 28 & \textbf{96} & 3 & 92 & \textbf{61} \\
    Grab Roller & 98 & 0 & 74 & 43 & 96 & 80 & 99 & 28 & 99 & 41 & 95 & 57 & 95 & 4 & \textbf{100} & \textbf{94} \\
    Handover Block & 10 & 0 & 45 & 14 & 45 & 8 & 4 & 0 & 38 & 0 & 81 & \textbf{36} & 5 & 0 & \textbf{83} & 19 \\
    Handover Mic & 53 & 0 & 90 & 31 & 98 & 13 & 45 & 0 & 75 & 8 & 96 & 61 & \textbf{100} & 0 & 99 & \textbf{86} \\
    Hanging Mug & 8 & 0 & 23 & \textbf{16} & 11 & 3 & 4 & 0 & 12 & 1 & \textbf{48} & 5 & 32 & 0 & 46 & \textbf{16} \\
    Lift Pot & 39 & 0 & 72 & 9 & 84 & 36 & 20 & 0 & 87 & 32 & 95 & 34 & 89 & 0 & \textbf{100} & \textbf{46} \\
    Move Can Pot & 39 & 0 & 25 & 12 & 58 & 21 & 48 & 0 & 78 & 0 & \textbf{94} & 15 & 85 & 7 & 87 & \textbf{24} \\
    Move Pillbottle Pad & 1 & 0 & 8 & 0 & 21 & 1 & 51 & 7 & \textbf{92} & 1 & 76 & 26 & \textbf{92} & 1 & 90 & \textbf{34} \\
    Move Playingcard Away & 47 & 0 & 43 & 11 & 53 & 22 & 79 & 13 & 92 & 30 & 89 & 53 & \textbf{99} & 2 & 98 & \textbf{89} \\
    Move Stapler Pad & 1 & 0 & 2 & 0 & 0 & 2 & 8 & 0 & 27 & 0 & \textbf{45} & \textbf{19} & 35 & 0 & 24 & 11 \\
    Open Laptop & 49 & 0 & 59 & 32 & 85 & 46 & 86 & 21 & \textbf{96} & 37 & 89 & 37 & 90 & 8 & 94 & \textbf{59} \\
    Open Microwave & 5 & 0 & 37 & 20 & 80 & \textbf{50} & 2 & 7 & 0 & 0 & \textbf{86} & 24 & 44 & 1 & 50 & 2 \\
    Pick Diverse Bottles & 6 & 0 & 2 & 0 & 27 & 6 & 52 & 18 & \textbf{83} & 34 & 76 & 17 & 69 & 4 & 81 & \textbf{50} \\
    Pick Dual Bottles & 24 & 0 & 42 & 13 & 57 & 12 & 82 & 31 & \textbf{93} & \textbf{56} & 82 & 30 & 76 & 9 & \textbf{93} & \textbf{56} \\
    Place A2B Left & 2 & 0 & 3 & 1 & 31 & 1 & 74 & 4 & 79 & 12 & 68 & 8 & 76 & 1 & \textbf{90} & \textbf{69} \\
    Place A2B Right & 13 & 0 & 1 & 1 & 27 & 6 & 56 & 1 & 81 & 11 & 52 & 9 & 84 & 3 & \textbf{87} & \textbf{68} \\
    Place Bread Basket & 14 & 0 & 10 & 2 & 17 & 4 & 63 & 20 & 90 & 29 & 90 & 26 & \textbf{96} & 4 & 92 & \textbf{71} \\
    Place Bread Skillet & 11 & 0 & 5 & 1 & 23 & 1 & 71 & 16 & \textbf{91} & 26 & 85 & 23 & 85 & 3 & 90 & \textbf{60} \\
    Place Burger Fries & 72 & 0 & 50 & 27 & 80 & 4 & 97 & 26 & \textbf{99} & 11 & \textbf{99} & 37 & 96 & 8 & 90 & \textbf{79} \\
    Place Can Basket & 18 & 0 & 19 & 6 & 41 & 5 & 20 & 0 & 63 & 0 & 68 & 34 & 58 & 0 & \textbf{80} & \textbf{44} \\
    Place Cans Plasticbox & 40 & 0 & 6 & 5 & 34 & 2 & 66 & 24 & 94 & 5 & 98 & 47 & 93 & 2 & \textbf{99} & \textbf{48} \\
    Place Container Plate & 41 & 0 & 78 & 17 & 88 & 45 & 86 & 48 & \textbf{100} & 58 & 96 & 22 & 98 & 17 & 96 & \textbf{86} \\
    Place Dual Shoes & 8 & 0 & 4 & 4 & 15 & 0 & 45 & 0 & \textbf{57} & 0 & 15 & 3 & 24 & 0 & 42 & \textbf{24} \\
    Place Empty Cup & 37 & 0 & 56 & 7 & 37 & 11 & 74 & 27 & 97 & 34 & 95 & 28 & \textbf{98} & 10 & 92 & \textbf{86} \\
    Place Fan & 3 & 0 & 12 & 2 & 20 & 10 & 31 & 1 & 62 & 5 & 62 & 9 & 75 & 1 & \textbf{76} & \textbf{42} \\
    Place Mouse Pad & 0 & 0 & 1 & 0 & 7 & 1 & 27 & 0 & 46 & 14 & 51 & 12 & 71 & 0 & \textbf{73} & \textbf{34} \\
    Place Object Basket & 15 & 0 & 33 & 17 & 16 & 2 & 56 & 1 & 66 & 3 & 89 & 25 & 50 & 2 & \textbf{93} & \textbf{78} \\
    Place Object Scale & 1 & 0 & 1 & 0 & 10 & 0 & 36 & 4 & 71 & 0 & 55 & 5 & 80 & 0 & \textbf{91} & \textbf{53} \\
    Place Object Stand & 22 & 0 & 15 & 5 & 36 & 11 & 76 & 24 & 87 & 21 & 84 & 33 & \textbf{95} & 12 & 93 & \textbf{72} \\
    Place Phone Stand & 13 & 0 & 15 & 6 & 35 & 7 & 32 & 0 & 61 & 2 & \textbf{91} & 10 & 85 & 0 & 79 & \textbf{52} \\
    Place Shoe & 23 & 0 & 35 & 7 & 28 & 6 & 76 & 12 & 90 & 29 & 70 & 18 & 83 & 7 & \textbf{91} & \textbf{79} \\
    Press Stapler & 6 & 0 & 41 & 24 & 62 & 29 & 79 & 56 & 94 & 58 & 92 & \textbf{64} & 58 & 19 & \textbf{99} & 45 \\
    Put Bottles Dustbin & 22 & 0 & 21 & 4 & 54 & 13 & 7 & 0 & 42 & 10 & 80 & 13 & 83 & 2 & \textbf{93} & \textbf{43} \\
    Put Object Cabinet & 42 & 0 & 33 & 18 & \textbf{68} & 18 & 7 & 0 & 52 & 0 & 59 & 8 & 41 & 0 & 45 & \textbf{25} \\
    Rotate QRcode & 13 & 0 & 50 & 5 & 68 & 15 & 56 & 2 & \textbf{81} & 21 & 69 & 11 & 76 & 0 & 79 & \textbf{40} \\
    Scan Object & 9 & 0 & 4 & 1 & 18 & 1 & 47 & 23 & 77 & 32 & 73 & 24 & 77 & 4 & \textbf{81} & \textbf{59} \\
    Shake Bottle Horizontally & 59 & 18 & 84 & 51 & 99 & 51 & \textbf{100} & 68 & \textbf{100} & 73 & \textbf{100} & 66 & \textbf{100} & 41 & \textbf{100} & \textbf{97} \\
    Shake Bottle & 65 & 8 & 74 & 45 & 97 & 60 & 98 & 54 & \textbf{100} & 74 & 98 & 73 & \textbf{100} & 49 & \textbf{100} & \textbf{98} \\
    Stack Blocks Three & 0 & 0 & 2 & 0 & 17 & 0 & 8 & 0 & 45 & 5 & 96 & 37 & 0 & 0 & \textbf{97} & \textbf{51} \\
    Stack Blocks Two & 7 & 0 & 21 & 2 & 42 & 1 & 61 & 0 & 95 & 6 & \textbf{100} & 60 & 2 & 0 & \textbf{100} & \textbf{75} \\
    Stack Bowls Three & 63 & 0 & 51 & 17 & 66 & 24 & 42 & 1 & 63 & 13 & \textbf{92} & 25 & 77 & 1 & 86 & \textbf{68} \\
    Stack Bowls Two & 61 & 0 & 76 & 30 & 91 & 41 & 69 & 12 & 90 & 52 & \textbf{98} & 49 & 96 & 11 & 90 & \textbf{81} \\
    Stamp Seal & 2 & 0 & 1 & 0 & 3 & 4 & 34 & 2 & \textbf{77} & 8 & 60 & 21 & 68 & 0 & 73 & \textbf{30} \\
    Turn Switch & 36 & 1 & 35 & 15 & 27 & 23 & 43 & 26 & 49 & 30 & \textbf{65} & 27 & 56 & 17 & 62 & \textbf{65} \\
    \midrule
    \rowcolor[HTML]{FFF0F5}
    Average & 28.0 & 0.6 & 34.5 & 13.7 & 46.4 & 16.3 & 52.9 & 15.2 & 75.3 & 20.5 & 80.5 & 26.4 & 71.9 & 6.3 & \textbf{84.0} & \textbf{55.7} \\
    \bottomrule
    \end{tabular}}
    \label{tab:exp_robotwin_total}
    \end{table*}

\paragraph{Quantitative results.}
Table~\ref{tab:exp_robotwin_preview} provides the complete per-task RoboTwin~2.0 results corresponding to the summarized benchmark table in the main paper. The full table is included to expose task-level variation that is hidden by the preview table and average success rates. For each task, we report both Clean and Random success rates for all compared VLA and WAM baselines, allowing readers to inspect whether the aggregate trends are consistent across different manipulation categories rather than driven by a small subset of tasks. The complete results show that performance varies substantially across task types. MV-WAM is especially strong on tasks involving visual attribute recognition, spatial placement, and multi-object stacking, such as \textit{Blocks Ranking RGB}, \textit{Place Object Basket}, \textit{Stack Blocks Three}, and \textit{Stack Blocks Two}. The table also reveals remaining difficult cases, including \textit{Open Microwave} and \textit{Move Stapler Pad}, where narrow affordances, contact precision, or unfavorable randomized layouts can still degrade execution. These task-level results complement the main-paper summary by identifying where the proposed model generalizes reliably and where future improvements in local affordance modeling and contact-sensitive control may be most useful.

\begin{figure}[t]
    \centering
    \includegraphics[width=0.99\textwidth]{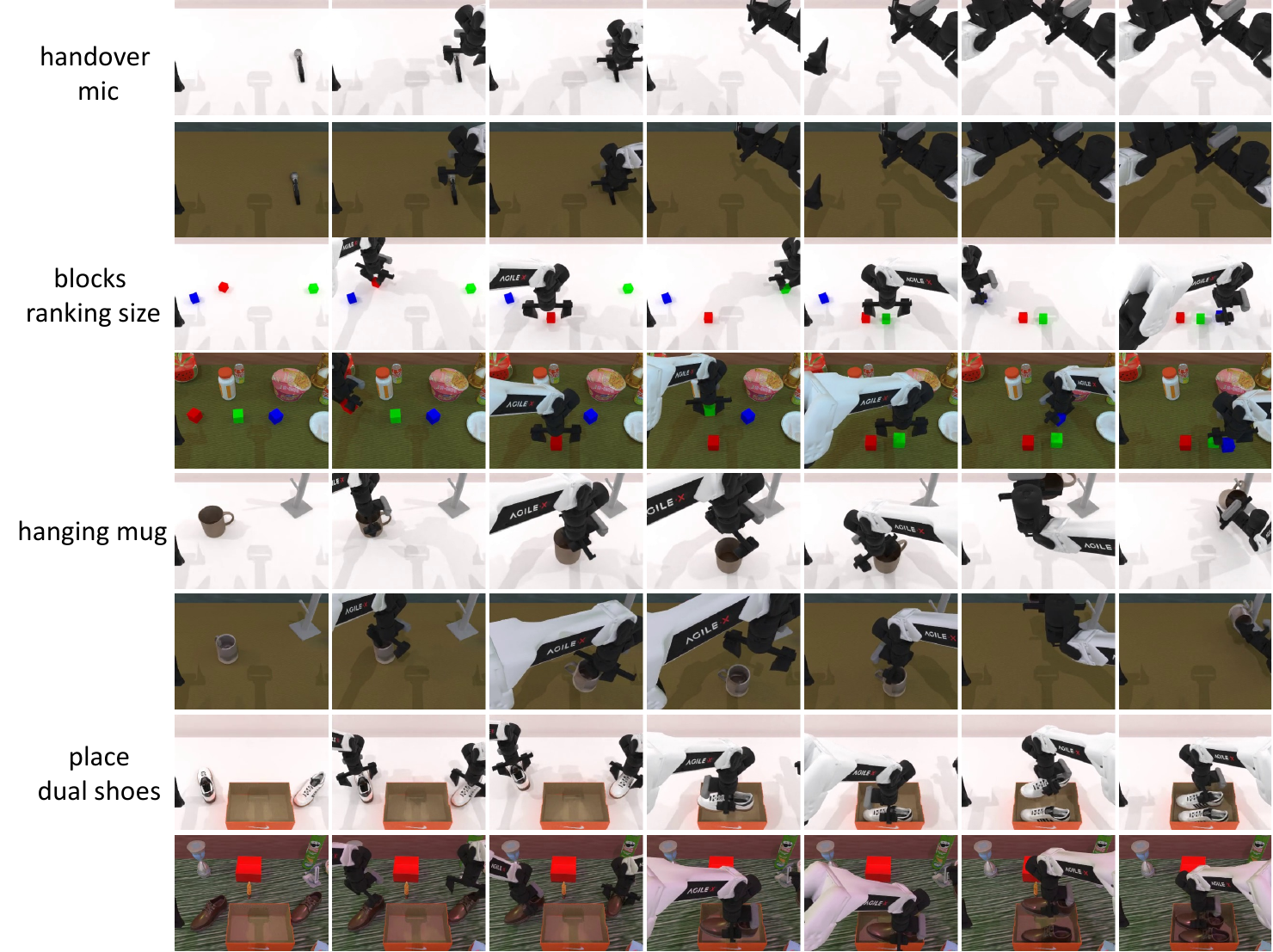}
    \caption{
        \textbf{Robot execution progress in RoboTwin~2.0 simulation tasks.} 
        We visualize key frames of the robot's execution process from a static exterior view in simulation tasks, including both clean and random scenes.
    }
    \label{fig:robotwin_execution_progress}
\end{figure}

\paragraph{Qualitative results.}
In addition to the quantitative results, we provide qualitative rollout visualizations in Fig.~\ref{fig:robotwin_execution_progress}. These examples illustrate the execution process of MV-WAM under diverse RoboTwin~2.0 tasks and randomized visual conditions. The qualitative results show that MV-WAM can generate temporally consistent action sequences while maintaining stable object interaction across different task categories, including object placement, stacking, and attribute-conditioned manipulation. Notably, the model is able to preserve coherent visual-action alignment under background changes, object layout variations, and distractor objects, which further supports the quantitative improvements observed in the Random setting.

\begin{figure}[ht]
    \centering
    \includegraphics[width=0.99\textwidth]{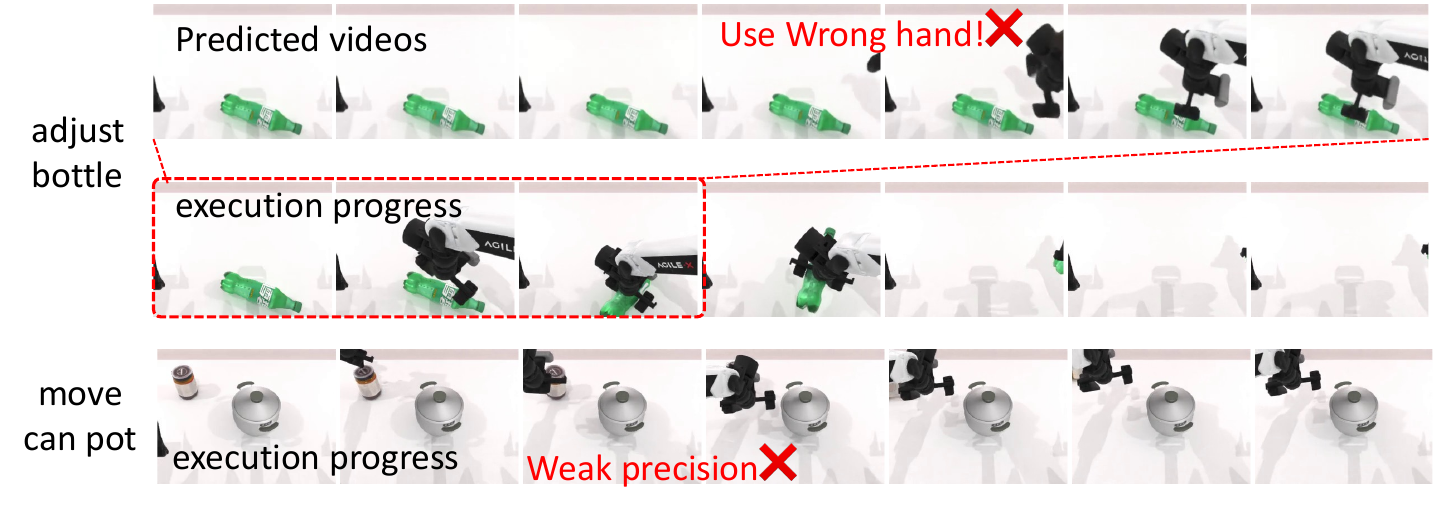}
    \caption{
        \textbf{Wrong Robot execution progress in RoboTwin2.0 simulation tasks.}
    }
    \label{fig:fail_analysis}
\end{figure}

\paragraph{Failure analysis.}
Although MV-WAM achieves strong overall generalization, we observe two representative failure modes in qualitative rollouts, as illustrated in Fig.~\ref{fig:fail_analysis}. The first type of failure is caused by the limited semantic and spatial understanding capacity of the Video Expert. For example, in the \textit{Adjust Bottle} case, the predicted visual rollout captures the general object motion but selects the wrong manipulator: the task should be executed with the left hand, whereas the model incorrectly drives the right hand. This suggests that the visual imagination branch can still produce plausible but semantically misaligned futures when the task requires fine-grained reasoning about arm assignment, object-side relations, or embodiment-specific constraints. The second type of failure comes from insufficient action precision in contact-sensitive scenarios. In the \textit{Move Can Pot} case, the model predicts a generally correct interaction direction, but the generated action trajectory lacks the fine-grained positional accuracy required for successful execution. Such errors often appear when the task involves small affordances, narrow contact regions, or precise end-effector alignment. These observations indicate that future improvements should strengthen both high-level visual understanding for embodiment-aware planning and low-level action accuracy for fine-grained contact control.

\section{Broader Impact}

MV-WAM studies more robust and generalizable robotic manipulation by improving how robot policies adapt to diverse visual conditions and execution scenarios. Such systems have the potential to improve productivity and quality of life by assisting with repetitive, physically demanding, or hazardous tasks in domains such as manufacturing, logistics, healthcare support, and domestic service. More generalizable manipulation policies may also reduce the engineering cost required to adapt robots to new environments, making robotic assistance more accessible across diverse real-world settings.

At the same time, deploying capable manipulation systems raises important safety and societal concerns. Robots operating in open environments may cause unintended physical damage or safety risks when facing unfamiliar objects, humans, or environmental conditions. Large-scale deployment could also affect labor markets by automating certain manual tasks and may introduce accountability challenges when failures occur. Therefore, systems such as MV-WAM should be deployed only with rigorous safety evaluation, human oversight, clear operational boundaries, and mechanisms for monitoring and intervention. Future work should continue to study robustness, failure modes, and safe human-robot interaction before real-world deployment.




\end{document}